\documentclass[preprint,journal]{vgtc}       % preprint (journal style)

\ifpdf%                                % if we use pdflatex
  \pdfoutput=1\relax                   % create PDFs from pdfLaTeX
  \usepackage{graphicx}                % allow us to embed graphics files
  \DeclareGraphicsExtensions{.pdf,.png,.jpg,.jpeg} % for pdflatex we expect .pdf, .png, or .jpg files
\else%                                 % else we use pure latex
  \usepackage{graphicx}                % allow us to embed graphics files
  \DeclareGraphicsExtensions{.eps}     % for pure latex we expect eps files
\fi%

\graphicspath{{figures/}{pictures/}{images/}{./}} % where to search for the images

\usepackage{microtype}                 % use micro-typography (slightly more compact, better to read)
\PassOptionsToPackage{warn}{textcomp}  % to address font issues with \textrightarrow
\usepackage{textcomp}                  % use better special symbols
\usepackage{mathptmx}                  % use matching math font
\usepackage{times}                     % we use Times as the main font
\usepackage{cite}                      % needed to automatically sort the references
\usepackage{tabu}                      % only used for the table example
\usepackage{booktabs}                  % only used for the table example
\usepackage{amsmath,amsfonts,amssymb}

\usepackage{xcolor}
\usepackage{xspace}
\usepackage{lipsum}
\usepackage{comment}
\usepackage{paralist}
\usepackage{hyperref}
\usepackage{enumerate}
\usepackage[inline]{enumitem}
\usepackage[noend]{algorithmic}
\usepackage{fontawesome}
\newcommand{\jianben}[1]{{\color{black} #1}}

\newcommand{\xingbo}[1]{{\color{black} #1}}
\newcommand{\xbRevise}[1]{{\color{black} #1}}

\newcommand{\ie}{i.e.}
\newcommand{\eg}{e.g.}
\newcommand{\etal}{et al.}
\newcommand{\imp}[1]{\textbf{\textit{{#1}}}}

\newcommand{\systemname}{{M\textsuperscript{2}Lens}}
\newcommand{\name}{{\textit{M\textsuperscript{2}Lens}}}
\newcommand{\vone}{{\textit{User Panel}}}
\newcommand{\vtwo}{{\textit{Summary View}}}
\newcommand{\vthree}{{\textit{Template View}}}
\newcommand{\vfour}{{\textit{Projection View}}}
\newcommand{\vfive}{{\textit{Instance View}}}

\usepackage{algorithm}
\usepackage{algorithmic}
\usepackage{lineno}
\newcommand{\fontsmall}{\fontsize{8pt}{10pt}\selectfont}
\newcommand{\qbox}[1]{%
	\medskip
	\fcolorbox{gray}{white}{
		\begin{minipage}{0.94\linewidth}
			\begin{internallinenumbers}
			\resetlinenumber
			\fontsmall
			\emph{#1}
			\end{internallinenumbers}
		\end{minipage}
	}
	\medskip
}

\onlineid{1112}

\vgtccategory{Research}
\vgtcpapertype{Analytics \& Decisions}

\title{{\systemname}: Visualizing and Explaining Multimodal Models for \\ Sentiment Analysis}

\author{Xingbo Wang, Jianben He, Zhihua Jin, Muqiao Yang, Yong Wang, and Huamin Qu}
\authorfooter{
\item
 Xingbo Wang, Jianben He, Zhihua Jin, and Huamin Qu are with the Hong Kong University of Science and Technology. E-mail: \{xingbo.wang, jhebt, zjinak, huamin\}@ust.hk.
\item
 Muqiao Yang is with Carnegie Mellon University. E-mail: muqiaoy@andrew.cmu.edu.
\item
 Yong Wang is with Singapore Management University. E-mail: yongwang@smu.edu.sg. He is the corresponding author.
}

\shortauthortitle{Biv \MakeLowercase{\textit{et al.}}: Global Illumination for Fun and Profit}
\abstract{Multimodal sentiment analysis aims to recognize people’s attitudes from
multiple communication channels such as verbal content (\ie, text), voice, and facial expressions. It has become a vibrant and important research topic in natural language processing. Much research focuses on modeling the complex intra- and inter-modal interactions between different communication channels.
However, current multimodal models with strong performance are often deep-learning-based techniques and work like black boxes.
It is not clear how models utilize multimodal information for sentiment predictions.
Despite recent advances in techniques for enhancing the explainability of machine learning models, they often target unimodal scenarios (\eg, images, sentences),
and little research has been done on explaining multimodal models.
In this paper, we present an interactive visual analytics system, {\name}, to visualize and explain multimodal models for sentiment analysis. {\name} provides explanations on intra- and inter-modal interactions at the global, subset, and local levels. Specifically, it summarizes the influence of three typical interaction types (\ie, dominance, complement, and conflict) on the model predictions. 
Moreover, {\name} identifies frequent and influential multimodal features and supports the multi-faceted exploration of model behaviors from 
language, acoustic, and visual modalities.
Through two case studies and expert interviews, we demonstrate our system can help users gain deep insights into the multimodal models for sentiment analysis.
}

\keywords{Multimodal models, sentiment analysis, explainable machine learning}
\CCScatlist{ % not used in journal version
 \CCScat{K.6.1}{Management of Computing and Information Systems}%
{Project and People Management}{Life Cycle};
 \CCScat{K.7.m}{The Computing Profession}{Miscellaneous}{Ethics}
}

\teaser{
  \centering
  \includegraphics[width=0.9\linewidth]{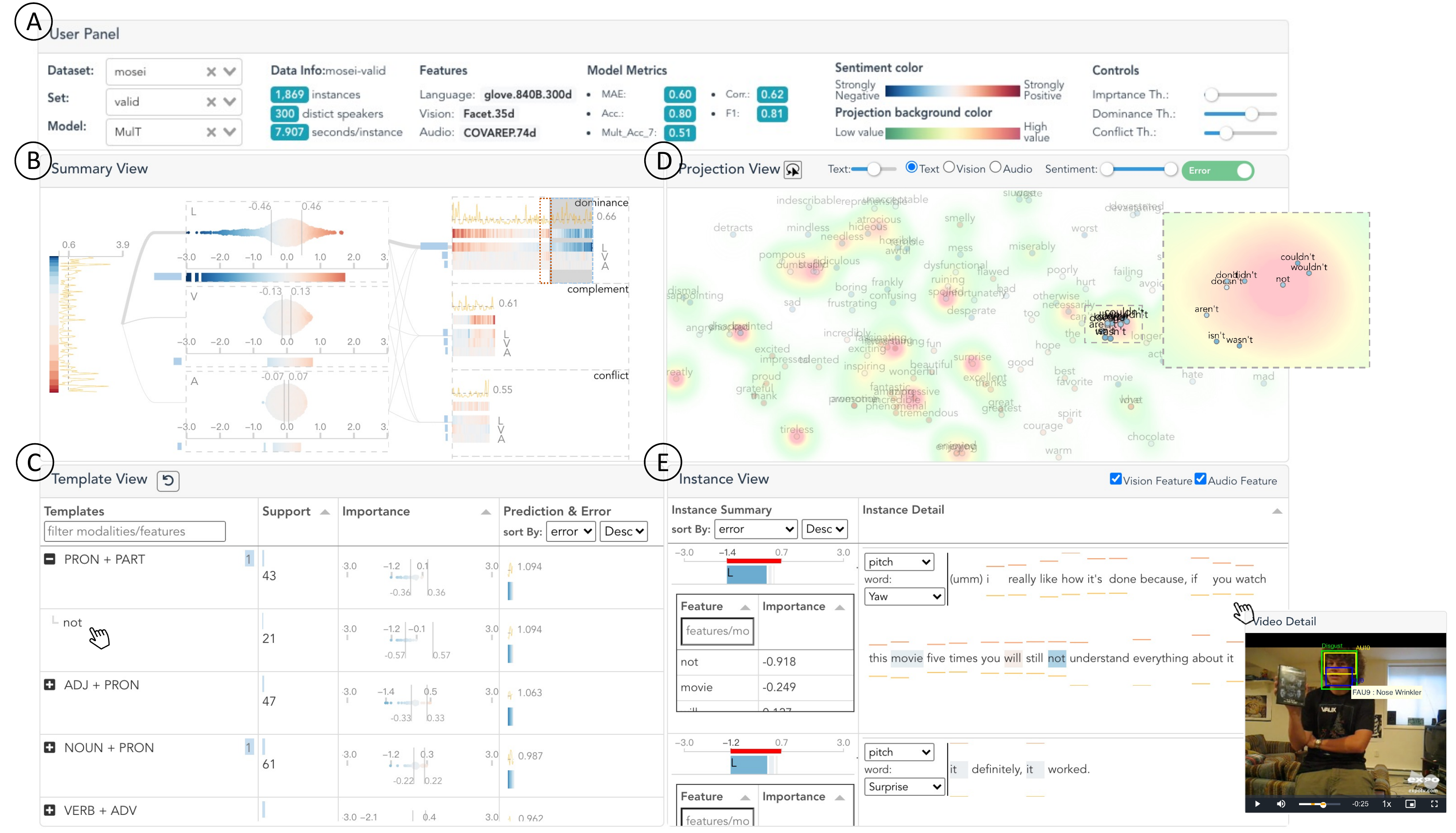}
  \vspace{-2mm}
  \caption{The explanatory interface of {\name} consists of five views. The {\vone} ~\xbRevise{(A)} displays the descriptive statistics about the model and dataset. The {\vtwo}~\xbRevise{(B)} presents a global summary of the importance of individual modalities, as well as their interactions using a three-layer augmented tree-like layout. The {\vthree}~\xbRevise{(C)} and {\vfour}~\xbRevise{(D)} complement each other for subset-level explanations. Specifically, {\vthree}~\xbRevise{(C)} summarizes frequent and influential templates of feature sets in a table. The {\vfour}~\xbRevise{(D)} supports multi-faceted explorations of instances that have features of interest. The {\vfive}~\xbRevise{(E)} provides local explanations by visualizing the important features and the context of individual instances.}
  \label{fig:teaser}
}

 \vgtcinsertpkg

\begin{document}


\firstsection{Introduction}
\maketitle

Sentiment analysis aims to 
\xingbo{use computational approaches to identify}
% use natural language processing and text analysis to identify
% affective states and 
people's attitudes, opinions, and other 
subjective information in human communication.
It can benefit various applications, such as customer analysis, social robots, and political campaigns.
Prior research on sentiment analysis is mainly based on \xingbo{a single communication channel} (i.e., text or facial expression)~\cite{zeng2008survey, sariyanidi2014automatic, soleymani2017survey}, which is often referred to as \textit{unimodal sentiment analysis}.
However, human communication is often multimodal.
For example, people can show their happiness through positive words and tones, along with a wild smile.
With the thriving of social media,
a large number of multimodal communication datasets can be collected and studied, e.g., TV series and vlogs \xingbo{showing} people's sentiment towards different topics and objects.
This has greatly boosted the development of \textit{multimodal sentiment analysis} techniques, and it has already become a vibrant and important research topic.

% multimodal (v.s. unimodal)

% \yong{Can we consistently use the term ``multimodal sentiment analysis'' and repalce all the terms ``multimodal language analysis''?}
Unlike the 
% \xingbo{
long-established
% } 
% and text-based
unimodal sentiment analysis, 
% multimodal language analysis 
multimodal sentiment analysis
combines the heterogeneous data and captures two primary forms of interactions in different modalities: \imp{intra-modal} and \imp{inter-modal} interactions. 
Intra-modal interactions refer to the dynamics of one modality, which 
% is similar to 
% \xingbo{
is the same as
the unimodal analysis based on the single communication channel.
% e.g., a spoken sentence or a series of facial expressions.
Inter-modal interactions consider the correspondence between different modalities across time, e.g., the co-occurrences of a happy tone and a smile or a sudden pause after a humorous punchline.
In practice, people's communication styles are highly complex and idiosyncratic.
For example, a sentence may seem semantically positive, but people can express it with a sarcastic tone to reveal their dissatisfaction.
In such cases,
unimodal sentiment analysis is not reliable, while multimodal models can offer the opportunities to explore vocal and visual expressions besides texts.
In addition, previous research~\cite{poria2017review, soleymani2017survey, baltruvsaitis2018multimodal} has confirmed that multimodal models are more accurate and robust in various downstream tasks.

% modal complexity and interpretability issues
% Recent research has proposed deep learning models, such as transformer, CNN, RNN, to fuse multimodal information with superior performance compared to traditional machine learning models.
Currently, deep-learning-based models achieve superior performance over the traditional methods~\cite{wiebe2005annotating, morency2011towards} in multimodal sentiment analysis.
% Recently, many deep-learning-based methods have been proposed to conduct multimodal language analysis.
% due to their excellent performance. 
% \xingbo{They achieve superior performance over the traditional methods~\cite{wiebe2005annotating, morency2011towards}.}
Representative examples include transformers~\cite{tsai2019multimodal, rahman2019integrating}, Convolutional Neural Networks (CNNs)~\cite{wang2017select}, and Recurrent Neural Networks (RNNs)~\cite{zadeh2018multi, rajagopalan2016extending, poria2017context}.
However, these models often work like black-boxes, 
hindering users from understanding the underlying model mechanism and fully trusting them in decision-making.
% The model explainability is essential for model developers to understand model decisions, discover model limitations, improve new designs, as well as diagnose the quality of dataset and feature engineering.
Enhancing the explainability of deep learning models has become critical for both model developers and users, and received increasing attention 
% in the \xingbo{machine learning}
% community 
in the past few years~\cite{hohman2018visual, arrieta2020explainable}.
% Although pos-hoc explainability techniques (\eg, LIME, SHAP, and IG) help extract the important explanatory features that influence the model predictions, they are usually applied to uinmodal scenarios (\eg, image classification, opinion mining), failing to adapt to the multimodal setting.
For example, post-hoc explainability techniques, such as LIME~\cite{ribeiro2016should}, SHAP~\cite{lundberg2017unified}, and IG~\cite{sundararajan2017axiomatic}, help identify important features (\eg, words or image patches) that influence model predictions. 
\xingbo{However, these methods often target providing local explanations on instances (\eg, sentences) in unimodal scenarios.
They do not scale well to produce global explanations on how intra- and inter-modal interactions influence the model decisions, for example, how the models will behave when positive words and sad voices are presented.}
% However, they are often designed for uinmodal scenarios (\eg, image classification, opinion mining), and cannot work well in multimodal settings where audio, images, and texts
% % the information of multiple channels (i.e., audio, image and text) 
% will be the model input.
% \yong{pls check my side comments.}
% It is challenging to explain multimodal models.

\xingbo{
% Explaining multimodal models is challenging.
It is challenging to explain multimodal models for sentiment analysis.}
% First, they do not explicitly explain the interplay of different modalities, while inter-modal interactions are unique characteristics of multimodal language data and tasks. 
% They can help the users discern speakers' communication styles and diagnose whether the model decisions reside in the proper behavioral cues.
\xingbo{First, 
it is necessary to relate the model performance back to the multimodal input data~\cite{patel2008investigating, ren2016squares}.
The heterogeneity and high dimensionality of multimodal human behaviors make
% the feature explanations less understandable to human.
it difficult for users to easily interpret the input features or data, as well as how they affect model decisions. Compact and human-friendly summaries of multimodal data are highly desired, but 
little research (if not no) has been done on it.}
% still remains less researched.
% Naively presenting an exhaustive list of explanations will overwhelm the users and make it difficult to track the temporal or contextual information in the multimodal language.
Second, it is non-trivial to explain inter-modal interactions between different modalities explicitly, which, however, are the unique characteristics of multimodal sentiment analysis models.
\xingbo{For example, when a person says something positive with a  neutral voice and facial emotion, users may feel interested in whether the models can discern positive sentiment in the language modality (i.e., the text).}
% enables users to decide whether model decisions reside in the proper behavioral cues.
% They can help the users discern speakers' communication styles and diagnose whether the model decisions reside in the proper behavioral cues.
% Third, the ways people express their emotions and sentiments are often
% % idiosyncratic and 
% vaguely-defined and can vary across different people. 
% % Exploratory analysis is needed to investigate data patterns and reflect on model designs.
% Model developers and users may want to investigate the characteristics of multimodal inputs and their influences on model performances.

% To better support the diagnosis of multimodal machine learning models and help gain a deep understanding of multimodal human language, 
In this paper,
we propose {\name}, a novel explanatory visual analytics tool
% (\autoref{fig:teaser}), 
% \yong{it is weird to refer to a figure here.}
to help \xbRevise{both developers and users of multimodal machine learning models}
% and natural language processing
better understand and diagnose \underline{M}ultimodal
\underline{M}odels for sentiment analysis.
% and help gain a deep understanding of multimodal human language. 
% While most existing visual analytics systems focus on the individual models and unimodal analysis,
\xingbo{
% Based on 
By considering
the feature importance measured by post-hoc explainability techniques, {\name} interprets intra- and inter-modal interactions learned by a multimodal language model from the global, subsets, and local levels. Particularly, we focus on interpreting three typical types of interactions, i.e., \imp{dominance}, \imp{complement}, and \imp{conflict}.
Moreover, it facilitates a multi-faceted exploration of the multimodal features and their influences on the model decisions for sentiment analysis. 
% Our explanatory interface 
{\name}
consists of four major views.
Specifically,
the {\vtwo} features an augmented tree-like layout for global explanations of the impacts of individual modalities and their interplay.
The {\vthree} summarizes influential and frequent multimodal features with compact templates.
The {\vfour} enables multi-faceted exploration of the features of user interest using different glyph designs.
% The {\vtwo} presents a overview of the importance of individual modalities, as well as their interactions.
% The {\vthree} summarizes multimodal feature sets that contribute significantly to the output with representative templates.
% The {\vfour} enables exploration of the instances with the features of interest from multiple aspects, including the language, vision, and audio modality.
The {\vfive} 
% explains multimodal instances in detail.}
visualizes individual multimodal instances and their explanations with details.}

% \begin{comment}
% s% 2. 
% {\name} facilitates systematic exploration of the modeling of intra-modal and inter-modal interactions for multimodal models and data from the global level, group level, to local level. 
% Moreover, it assists in comparing a multimodal model with its ablation baselines to reveal the benefits that multimodality brings to the analysis.
% Specifically, our explanatory interface consists of five major views. The {\vone} shows the overall statistics of the dataset, and a multimodal model, along with its ablation baselines. It offers a quantitative measure of to what extent multimodal analysis outperforms the unimodal and bimodal analysis.
% For more detailed exploration, 
% The {\vtwo} presents a visual summary of the modality relevance to the model decisions. Furthermore, it explicitly reveals the interplay among different modalities.
% The {\vthree} summarizes groups of multimodal language styles with representative templates.
% The {\vfour} facilitates multi-facet exploration of multimodal features.
% The {\vfive} visualizes individual multimodal instances with the most detail.

In summary, our major contributions are:
\begin{compactitem}
    \item {\name}, a visual analytics system to produce multi-level and multi-faceted explanations on intra- and inter-modal interactions that are learned by the multimodal models.
    % support interactive and scalable interpretations and exploration of multimodal model behavior. % and multimodal language data.
    \item Case studies and expert interviews that demonstrate
    % our system 
    the effectiveness of our approach in
    helping users gain deep insights into two state-of-the-art multimodal models for sentiment analysis.
\end{compactitem}

\section{Related Work}
This section discusses the relevant research of our approach, including multimodal language analysis, post-hoc explainability techniques, and machine learning interpretation with visualization.

\subsection{Multimodal Sentiment Analysis}
\xingbo{
Multimodal sentiment analysis is a vibrant topic in natural language processing (NLP).
% and affective computing.
% It focuses on the automatic detection of 
It automatically extract people's attitudes or affective states
from multiple communication channels (\eg, text, voice, and facial expressions). 
Moreover, it has various applications \cite{zeng2019emoco, zeng2020emotioncues, hu2018multimodal}.
% The existing multimodal datasets are mostly opinion videos, such as movie review~\cite{perez2013utterance, zadeh2016multimodal}, product review~\cite{morency2011towards}, and dialogues~\cite{busso2008iemocap}.
% % \yong{Not sure if ``opinion video'' is a correct term.}
The core challenge 
% of multimodal sentiment analysis 
is modeling the complex \emph{intra-modal} and \emph{inter-modal} interactions, where multimodal features are being fused.}

Early work~\cite{ngiam2011multimodal, lazaridou2015combining} concatenated features from different modalities before being input to a learning model.
% for prediction. 
Conversely, some work adopted \emph{late-fusion} approaches that combine the decision values from individual unimodal models using a voting scheme
~\cite{potamianos2003recent, nojavanasghari2016deep} or a learning model~\cite{glodek2011multiple, ramirez2011modeling}. However, these methods ignore the cross-modal interactions. To address such issues, some work explicitly computed the unimodal, bimodal, and trimodal features and fused them with tensor product~\cite{zadeh2017tensor, liu2018efficient} and dynamic routing~\cite{tsai2020multimodal}.
Recently, neural network methods~\cite{zadeh2018multi , pham2019found, rajagopalan2016extending, chen2017multimodal, zadeh2018memory, tsai2019multimodal} are popular to model the complex interplay between modalities.
For example,
researchers~\cite{rajagopalan2016extending, chen2017multimodal} have extended LSTM cells and gates to learn temporal interaction patterns among multimodal sequences.
Pham \etal~\cite{pham2019found} proposed attention-based RNNs to learn multimodal representations with a cyclic translation loss among modalities.
Zadeh \etal~\cite{zadeh2018memory} designed a multi-view gated memory unit that is controlled by neural networks. It stores and predicts temporal cross-modal interactions.
Tsai \etal~\cite{tsai2019multimodal} utilized transformer attention mechanisms to learn both cross-modal alignment and interactions.
Although neural networks greatly improve the performance over traditional methods, their complex architecture seriously affects the model interpretability.
This paper presents an explanatory interface to diagnose black-box models for sentiment analysis tasks.

\subsection{Post-hoc Explainability Techniques}
Post-hoc explainability techniques 
interpret models after the training process
% refer to separate interpretation methods for model understanding after the model training process
\cite{arrieta2020explainable, longo2020explainable}. 
They generally include \emph{model-specific} and \emph{model-agnostic} approaches~\cite{longo2020explainable}. Model-specific methods explain particular models ranging from shallow models~\cite{tolomei2017interpretable,haasdonk2005feature} to sophisticated neural networks~\cite{krakovna2016increasing, selvaraju2017grad}.
% They can be roughly categorized into two groups:
% \emph{model-specific} and \emph{model-agnostic}~\cite{longo2020explainable} approaches. 
% Model-specific approaches 
% provide explanations for particular types of models ranging
% % are usually tailed to specific types of model explanation 
% % are specifically-designed explanations for different models ranging
% from shallow models like tree ensembles~\cite{tolomei2017interpretable, hara2018making} and support vector machines~\cite{haasdonk2005feature, barakat2007rule} to more sophisticated neural networks~\cite{krakovna2016increasing, selvaraju2017grad}.
In contrast, model-agnostic methods are flexible enough to be 
applied
to any machine learning model. 
Here, we discuss two main types of model-agnostic approaches: explanation by simplification and feature relevance explanation~\cite{arrieta2020explainable}.
For simplification techniques,
researchers often built surrogate models (\eg, 
rule-based learners~\cite{4734030, johansson2004accuracy, ribeiro2018anchors}, 
decision trees~\cite{bastani2017interpretability}, and linear models~\cite{ribeiro2016should}) to imitate the original model behaviors with reduced complexity.
% \yong{Pls do check if you use the correct term, I think it should be called "surrogate models": https://christophm.github.io/interpretable-ml-book/global.html}
% almost all techniques are rule-extraction based where a new simplified model or system with similar performance scores is usually built to imitate the model behaviors with reduced complexity. Inspired by G-Rex\cite{4734030} which is initially brought up to extract rules from data, several work\cite{johansson2004accuracy}\cite{johansson2004truth} extend this practice to opaque model interpretation. Other simplification techniques include building linear models or decision trees based on model approximation to gain insights\cite{bastani2017interpretability} \cite{ribeiro2016should} and implementing distillation on black-box models to inspect the needed information to achieve similar prediction accuracy or model recreation\cite{tan2018distill}. 
% Local explanations identify significant and less complex input dimensions that contribute to a model decision~\cite{ribeiro2016should, wachter2017counterfactual, ribeiro2018anchors}.
One of the most representative methods is LIME~\cite{ribeiro2016should}, which builds locally linear models to approximate individual predictions based on neighbors of instances of interest.
% Among local explanations, one of the most prevalent technique is LIME~\cite{ribeiro2016should} which aims to train high-approximate local surrogate models on the samples of interest. Based on LIME, the author later proposed anchors\cite{ribeiro2018anchors}, a technique similarly focuses on local explanations for black-box models but adopts perturbation-based strategy and "if-then" rules for better result explanation instead of the surrogate models used by LIME. \jianben{advantages and disadvantages of LIME and anchors?}. Another different type of local explanation approaches is counterfactual explanations. Counterfactual explanations describe the situations where subtle feature value changes led to different outputs of individual instances without looking into the full logic of the model. In\cite{wachter2017counterfactual}, by comparing the result-changed instances with the original ones, the method identifies the minimal changes required to alter the decision. \jianben{counterfactual need your own words paraphrase}. 
Feature relevance explanation quantifies the feature contributions 
to model predictions.
% by computing the relevance or importance scores.
% of each feature to quantify their contributions to the to-be explained models. 
One popular example is SHAP~\cite{lundberg2017unified}, whose mathematical root is Shapley Value~\cite{shapley1953value}---a method from cooperative game theory. 
SHAP computes an additive importance score for each feature 
to describe its influence, 
given a prediction result. It has desirable properties (local accuracy, missingness, and consistency) and is proved to be aligned with human intuitions.
Other work used local gradients~\cite{robnik2008explaining}, randomized feature permutations~\cite{henelius2014peek}, or influence functions~\cite{koh2017understanding} to disclose feature relevance.
% to the model predictions.
% SHAP leverages the contribution of each feature on predictions to explain the models where shapley values tell us how to fairly distribute the predictions among the features.
% \jianben{say more about SHAP, its advantages? also the sentence "SHAP leverages...need you own word paraphrase}Similarly inspired by game theory, the authors in \cite{strumbelj2010efficient} proposed to replace features with random quantity to measure the influence between inputs and outputs. Apart from these game theory based methods, there are many other techniques which utilize local gradients test\cite{robnik2008explaining} or randomized feature permutations\cite{henelius2014peek} to unveil the importance of different features to models. Feature relevance explanation techniques also include some techniques based on sensitivity analysis\cite{cortez2011opening}\cite{cortez2013using} or influence functions\cite{koh2017understanding}\cite{johansson2004accuracy} to disclose the influence of evaluated features on the predicted output. 
% \jianben{there is no clear comparison, pros\&cons analysis, the reason to choose, for these 4 types of explanations, which may need you to add}

% Our system accepts multimodal data with a relevance score for each feature. Then, it supports interactive visual exploration of multimodal behaviors and their influence to the sentiment predictions from global level to local level.
% 1. shap
% 2. data + feature relevance scores
\xingbo{However, the methods above are often used to interpret specific instances of one modality (\eg, sentences, images), which cannot be directly applied to multimodal sentiment analysis.
This paper aims to fill the gap by enabling multi-level explanations on the learned intra- and inter-modal interactions from global, subsets, and local levels.}

\subsection{Machine Learning Interpretation With Visualization}
With the increasing complexity of both data and machine learning models, various visual analytics systems have been proposed to assist in understanding the model behaviors.
Besides measuring the model performance with computational metrics, users also need to explore when and why a model makes specific decisions~\cite{hohman2018visual}.
% so that they can develop trust in models, discover model limitations, and further refine model designs.
One of the most common and important interpretation strategies in previous work is to reveal the relationship between the \textit{input data} and \textit{model predictions}~\cite{hohman2018visual, arrieta2020explainable}. They can be categorized into two groups: \emph{instance exploration} and \emph{feature \& subset exploration}.

Instance visualization shows model behavior towards individual data samples.
Amershi \etal~\cite{amershi2015modeltracker} presented ModelTracker to support performance debugging with a visual summary of binary classification instances.
Ren \etal~\cite{ren2016squares} extended the performance visualization to multi-class scenarios with aligned vertical axis designs, while Kahng \etal~\cite{kahng2017cti} and Alsallakh \etal~\cite{bilal2017convolutional} adopted a matrix-like design for instance summary.
Apart from visualizing instance distributions, 
Kulesza \etal~\cite{kulesza2015principles} built an exploratory debugging prototype to enable users to explain corrections back to models.
In addition, there are tools~\cite{harley2015interactive, smilkov2017direct} that allow users to interactively probe models with provided inputs.
% interactions with user-provided inputs. 

Feature and subset visualization investigates how to surface the patterns groups of features~\cite{krause2014infuse, brooks2015featureinsight, krause2016interacting} and instances~\cite{zhang2018manifold, ahn2019fairsight, cabrera2019fairvis, wexler2019if} that affect model decisions. 
Brooks \etal~\cite{brooks2015featureinsight} developed FeatureInsight, which supports the feature ideation process with a visual summary of set errors.
Krause \etal~\cite{krause2014infuse} enabled exploration of the predictive power of feature candidates across different feature selection algorithms.
For specific applications in CV and NLP, features are often visualized as
image patches~\cite{olah2017feature, selvaraju2017grad, springenberg2014striving} or text segments~\cite{karpathy2015visualizing, gehrmann2019gltr}.
Besides, researchers built interactive tools to facilitate group-level exploration. Zhang \etal~\cite{zhang2018manifold} conducted feature attribution comparisons to inspect discrepancies across different data subsets. 
Some work~\cite{ahn2019fairsight, cabrera2019fairvis, wexler2019if} used fairness metrics to partition data into groups for model diagnosis.
% Rules~\cite{ming2018rulematrix} and prototypes~\cite{ming2019protosteer} can also be used to summarize group characteristics.

\xingbo{However, these methods do not consider exploring multimodal features
% , as well as 
and
determining how much they affect model decisions.
Our system facilitates multi-faceted exploration of multimodal features and generates multi-level visual explanations on their influences.}

% In our work, we categorize and summarize data instances based on the modality importance to the model predictions. In addition, we use frequent multimodal templates, as well as a suite of glyphs, to conduct multi-facet exploration of features and subsets.

\section{Background}
\label{sec:background}

% Before introducing our visual analytics system, we provide 
% This section introduces
% the related background of multimodal sentiment analysis, including multimodal dataset, feature engineering techniques, performance metrics, and intra- and inter-modal interactions.
\xbRevise{In multimodal sentiment analysis, 
a machine learning model predicts sentiment based on the visual, acoustic, and language features extracted from the raw video data.
This section introduces the related background about multimodal datasets, feature engineering techniques, performance metrics, and intra- and inter-modal interactions.}

% Each modality in multimodal sentiment analysis contributes to the result of sentiment prediction. For example, facial expressions in vision modality will be important to understand whether the speaker is positive or negative about his descriptions. Acoustic features will also help to determine the sentiment of the speaker, considering the speaker may change the volume or pitch when expressing his or her emotions. Language modality, containing the words that the speaker uses to convey opinions, is also an important source of information to judge the sentiment of the speaker's statement.

\subsection{Dataset}
\label{subsec: dataset}

There is a wide range of multimodal datasets in the community. For example, \textbf{IEMOCAP}~\cite{busso2008iemocap} contains 151 videos of dialogues with different emotion labels. \textbf{YouTube}~\cite{morency2011towards} consists of videos of product reviews extracted from the social media website, YouTube. Without loss of generality, 
our work focuses on the largest and widely-used benchmark dataset for multimodal sentiment analysis, i.e.,
% our work will be focused on the multimodal sentiment analysis of the benchmark dataset,
\textbf{CMU-MOSEI}~\cite{zadeh2018multimodal}.
% , since it contains all three modalitites (vision, audio and language) and highest number of video hours. 
It consists of 23,454 monologue movie review video clips from 1,000 speakers and 250 topics in YouTube.
% , spanning over a wide range of speakers and topics. 
The sentiment of each video clip is labeled by three annotators with a Likert scale of $[-3, 3]$,
% assigned with a integer sentiment score ranging in $[-3, 3]$, 
where $3$ indicates strongly positive, $-3$ represents strongly negative, and $0$ means neural.
% $0$ means that this movie review is labeled as neutral. 
Besides the sentiment label, each video is associated with the information from the three communicative channels---transcripts for language resources ($l$), facial expressions for the visual ($v$), and voice of speakers as the acoustic modalities ($a$).

% Besides videos as the visual features, sound tracks as acoustic and transcripts as textual features are also associated with each movie review correspondingly, which make the three modalities in the dataset: vision ($v$), audio ($a$), and language ($l$). 
% Note that audio modality and vision modality features are extracted at a sampling rate at 20Hz and 15Hz respectively.

\subsection{Multimodal Feature Engineering}
\label{subsec: multimodal feature engineering}

% \xbRevise{Prior research mostly applies different feature engineering techniques to extract visual, acoustic, and language features from the raw data. Then, a model processes the multimodal features to make predictions.}
% \xbRevise{Prior research on multimodal sentiment an often applies different feature engineering techniques
% to compute mutlimodal features from the raw data.}
Prior research on multimodal models mostly uses different feature engineering techniques for all three modalities in sentiment analysis.
% \yong{Prior research on multimodal models?}
% for all three modalities in sentiment analysis.
Here, we follow the common practice of multimodal feature extraction (also provided by CMU-MOSEI).
For language features, transcripts are encoded by high-dimensional word vectors. We leverage Glove embeddings~\cite{pennington2014glove} to represent each word, where each word is transformed to a 300-dimension vector.
For visual modality, most work focuses on
% specific 
facial expressions, which are often encoded by Facial Action Coding System (FACS)
% \footnote{\url{https://en.wikipedia.org/wiki/Facial_Action_Coding_System}}
~\cite{friesen1978facial}. 
% \yong{Not understand.}
FACS encodes the facial muscle movement with 35 facial action units. We deploy it to extract frame-level facial features.
The
% low-level 
acoustic features are engineered through a speech processing framework, COVAREP~\cite{degottex2014covarep}. The extracted features have 74 dimensions, and all of them are related to speech emotions and tones.
% We include the details of these multimodal features in the supplementary material.
To help users 
% understand 
gain a quick overview of 
these 
% low-level 
fundamental
features, 
we further group them into 
% higher-level 
% \yong{why emphasizing ``high-level''?}
different
classes, which will be introduced in~\autoref{subsubsec: feature templates}.

% Here, we follow the common practice of multimodal feature extraction (also provided by CMU-MOSEI). Specifically, we use the Glove word embeddings~\cite{pennington2014glove}.
% 

% We introduce different feature engineering techniques in all three modalities to make our sentiment analysis and the embedding view more intuitive. For language modality, we use Glove word embeddings~\cite{pennington2014glove} for the transcripts, in which each word is represented by a 300-dimensional vector. For vision modality, we follow Facial Action Coding System (FACS)~\cite{friesen1978facial} to encode 35 facial action units in a frame-level. For audio modality, we use COVAREP~\cite{degottex2014covarep} to represent the acoustic features in four dimensions: amplitude, pitch, glottal, and phase. Visualization of the feature embeddings from different modalities will be included in Section \ref{sec:embeddingview}.

\subsection{Metrics for Multimodal Sentiment Analysis}
Prior work applies several metrics to evaluate the model performance for multimodal sentiment analysis, including
mean absolute error ($MAE$), the correlation between the model predictions and human labels ($Corr.$), F1 score ($F1$), 7-class accuracy ($Acc_7$), and 2-class accuracy ($Acc$). Note that $Acc_7$ considers all of the sentiment scores $\mathbb{Z} \in [-3, 3]$, while $Acc$ is a binary classification score that only predicts whether this video clip is positive or negative. 
% Comparisons between models in different metrics are shown in Fig \ref{fig:teaser}. 

% Following prior work~\cite{tsai2019multimodal}, we will illustrate our multimodal sentiment analysis in various metrics: 
% Mean Absolute Error (MAE), correlation between the model predictions and human labels ($Corr.$), 7-class accuracy ($Acc_7$), 5-class accuracy ($Acc_5$), F1 score ($F1$) and 2-class accuracy ($Acc$). Note that $Acc_7$ considers all of the sentiment scores $\mathbb{Z} \in [-3, 3]$, while $Acc_5$ removes the extreme sentiments ($-3$ and $3$) and $Acc$ is a binary classification score which only predicts whether this video clip is positive or negative. Comparisons between models in different metrics are shown in Fig \ref{fig:teaser}. 

\subsection{Intra- and  Inter-modal Interactions}
\label{subsec:modality-interaction}
In practice, sentiment
% expression often contains 
analysis relies on
multimodal language signals (\eg, language, facial expressions, and tones). 
A successful multimodal sentiment analysis requires the understanding of the combinations of these signals, where two primary forms of interactions exist---\imp{intra- and inter-modal} interactions~\cite{soleymani2017survey, baltruvsaitis2018multimodal}.

When modeling intra- and inter-modal interactions, three typical situations arise~\cite{soleymani2017survey, baltruvsaitis2018multimodal, zadeh2018memory}:
\begin{compactitem}
\item One modality is \imp{dominant} for sentiment analysis. For example, people may show agreement by nodding their heads, where the vision modality 
% determines 
dominantly indicates
their positive attitudes.
\item More than one modalities \imp{complement} each other when people are expressing their sentiment. 
For example, people's positive attitudes in words can be enhanced by a happy tone.
\item More than one modalities \imp{conflict} with each other.
% when expressing sentiment.
For example, people may tell sad stories with smiles on their face.
\end{compactitem}
Researchers have tried to build models to analyze the situations above for better sentiment analysis. However, 
most state-of-the-art models are deep-learning-based techniques with little interpretability. 
Model developers and users are not aware of how exactly the model utilizes information in multiple modalities in situations of dominance, complement, or conflict.
% happen. 
Explaining multimodal model behaviors not only provides insights into the multimodal language characteristics, but also reveals the model errors and inspires new model designs.
In our work, we explicitly provide global explanations on intra- and inter-modal interactions with a compact visual summary. Specifically, we categorize instances into dominance, complement, and conflict groups based on 
the importance of each modality 
computed by SHAP~\cite{lundberg2017unified}. 
% \yong{Do we explain ``instances'' when it first appears in the previous sections?}
Furthermore, we summarize influential feature sets for each group with templates to provide finer-grained explanations on model behavior.
\section{Design Requirements}
\label{sec:design_requirements}

% \yong{Task: need to check it.}

% Given key features of multimodal sentiment analysis (\autoref{sec:background}) and challenges for explaining multimodal models,
Our goal is to develop a visual analytics system to help users (\eg, model developers and model users) understand and diagnose the behaviors of multimodal models for sentiment analysis.
Similar to the general black-box explanation tools~\cite{amershi2015modeltracker, ren2016squares, krause2016interacting, zhang2018manifold}, 
interpreting multimodal models helps target users gain insights into the connection between the model performance (\eg, model errors) and the characteristics of multimodal data.
\xbRevise{For example, model users can examine whether a model has a bias or poor performance on some types of data and further decide if it is a proper fit for target applications.}
Furthermore, given the critical aspects of multimodal sentiment analysis (in \autoref{subsec:modality-interaction}),
it is beneficial to explain the intra- and inter-modal relationships learned by the model.
\xbRevise{For instance, model developers can adjust the fusion weights of different modalities based on their relative importance to achieve better sentiment predictions.
% impacts of the modalities.
However, it is challenging to interpret multimodal models due to the high complexity of multimodal data and inter-modal relationships.}

\xbRevise{
To understand users' general needs and formulate design requirements,
we surveyed prior visualization techniques for interpreting machine learning models
% ~\cite{hohman2018visual, arrieta2020explainable,liu2016towards,strobelt2017lstmvis,ming2017understanding,wang2018ganviz,wang2018dqnviz,jin2020gnnvis}
\cite{brooks2015featureinsight,amershi2015modeltracker,ren2016squares,krause2016interacting,krause2014infuse,zhang2018manifold,kahng2017cti,molnar2019,carvalho2019machine,arrieta2020explainable}
and multimodal language analysis~\cite{baltruvsaitis2018multimodal, ngiam2011multimodal,tsai2019multimodal,tsai2020multimodal,zadeh2018multimodal,zadeh2017tensor}.}
% Furthermore, we detailed and refined them by iteratively collecting feedback from a researcher in NLP and multimodal machine learning (who are also a co-author of this paper).
Also, we worked closely with a researcher in NLP and multimodal machine learning (who is also a co-author of this paper) for about five months to collect his feedback and iteratively refine the design requirements.
% detailed and refined them by iteratively collecting feedback from a researcher in NLP and multimodal machine learning (who are also a co-author of this paper).
% \yong{pls double-check if it is ``a researcher'' or ``researchers'' and if it is 5 months.}
We summarize the design requirements as follows.

\textbf{R1: Show the model performance.}
Performance metrics are crucial for guiding the model analysis~\cite{amershi2015modeltracker, ren2016squares}. 
They provide quantitative measures of how accurate the predictions are and can help users pinpoint where the model is likely to fail. The users often want to evaluate models at different levels:

\begin{compactitem}[]%\listparindent=3em\labelsep=6em\itemindent=3em
\setlength\labelsep{3em}
\item \imp{Q1:} \textit{What are the overall error distributions for model predictions?}
\item \imp{Q2}: \textit{What are the instances that are predicted with large/small errors?}
\end{compactitem}

\textbf{R2: Reveal the contributions of modalities to the model predictions.}
Besides performance metrics, the system should provide global explanations on how the model generally works, especially when working with huge datasets~\cite{kahng2017cti, molnar2019, carvalho2019machine, arrieta2020explainable}.
In multimodal sentiment analysis, intra- and inter-modal interactions are 
% the keys to understand 
crucial for understanding
the model behaviors\xbRevise{~\cite{baltruvsaitis2018multimodal, ngiam2011multimodal}.}
Thus, it is essential to summarize the influences of individual modalities and their interplay for predictions. Specifically, the system should help users answer the following questions:

\begin{compactitem}[]
\item \textit{\textbf{Q3}: How does each modality influence the model predictions?}
Displaying the contributions of each modality helps users prioritize their efforts in diagnosing a particular modality for model predictions\xbRevise{~\cite{zadeh2017tensor}}.

\item \textit{\textbf{Q4}: Which modalities \imp{dominate} the model predictions? Also, which modalities \imp{complement} or \imp{conflict} with each other for model predictions?}
To better reveal the characteristics of multimodal interactions captured by the model, 
the system should further summarize the instances according to the interaction types\xbRevise{~\cite{zadeh2018multimodal, tsai2020multimodal, tsai2019multimodal}.} 
Specifically, 
dominant, complementary, and conflicting modalities, which depict typical interaction types, are the targets for analysis.
% \yong{The last sentence needs further clarification.}

\item \textit{\textbf{Q5}: How do \imp{dominant/complementary/conflicting} modalities influence the model predictions?}
Besides recognizing the learned interaction types, it is also essential to connect them to the model predictions for a comprehensive understanding of model behaviors~\xingbo{\cite{amershi2015modeltracker, ren2016squares, zhang2018manifold}}.
For example, the dominance of language modality can contribute to positive or negative sentiment for different instances.
\end{compactitem}

% \yong{If possible, can we add some references to the above questions?}

\textbf{R3: Identify the influences of multimodal features for the model predictions.}
With a global understanding of how the model work on individual modality (\imp{R2}), users need to drill down to finer-level inspection on model behaviors. Feature-based exploration is a common and effective approach for explaining machine learning models~\cite{brooks2015featureinsight,krause2016interacting, krause2014infuse}. Accordingly, the system should connect high-level modality interactions with the corresponding multimodal features. 
For example, users may want to know when the language modality dominates the predictions and what words people use to express their sentiments.

\begin{compactitem}[]
\item \textit{\textbf{Q6}: What are the feature sets that significantly contribute to positive/negative sentiment predictions?}
Exploring all the features of instances individually is tedious given the high volume and dimensionality of multimodal data. 
Summarizing the set of features with a significant predictive contribution helps reduce the efforts in exploration\xbRevise{~\cite{brooks2015featureinsight, krause2014infuse}.}
% \yong{pls check if I changed your original idea.} 
In addition, it helps users develop a high-level concept about model predictions. For example, users may want to know what types of words or facial expressions are considered important to models when dealing with positive sentiment cases.
\item \textit{\textbf{Q7}: 
% For a instance, 
What features are considered important by the model? Are they plausible for prediction?}
To help users analyze the individual predictions, features
% that significant influences the model 
with a significant influence on the model performance
should be presented to users and allow them to judge whether they align well with the observation of the original data.
\end{compactitem}

\textbf{R4: Support multi-level and multi-faceted exploration of the multimodal model behaviors.}
% \xingbo{Xingbo: Revise here}
Given the multimodal settings of sentiment analysis, 
the visualization should empower users to explore the relationships between the model and input data from multiple aspects (\eg, language, facial expressions).
To facilitate a comprehensive understanding of multimodal models, explanations should be offered on different levels, including the influences of individual modalities and their interplay, and the importance of multimodal features.
% Furthermore, according to \imp{R1-R3}, users need to develop a comprehensive understanding of multimodal models on different levels---from global level (\imp{Q1}, \imp{R2}), subset level (\imp{Q6}), to local level (\imp{Q2}, \imp{Q7}).
% \yong{It is not a good idea to refer other requirements here.}
\section{\texorpdfstring{M\textsuperscript{2}Lens}{M^2Lens}}
\label{sec: system}

% \xingbo{Zhihua needs to check}

Based on the derived design requirements (\autoref{sec:design_requirements}), we develop a visual analytics system, {\name} (\autoref{fig:teaser}), for understanding and diagnosing how models utilize multimodal information for sentiment prediction. 
% It generates global, subset, and local level explanations and facilitates multi-faceted exploration of model behavior
In this section, we first provide an overview of the system architecture. Then, we will illustrate the methods for generating explanations of multimodal model behavior.
Next, we describe the visual designs and interactions in detail.
% visual designs and interactions in detail. When introducing our system and its designs, we will explain how they help address the design requirements.

\subsection{System Overview}
\begin{figure}[htb]
\vspace{-2mm}
\centering
\includegraphics[width=0.5\textwidth]{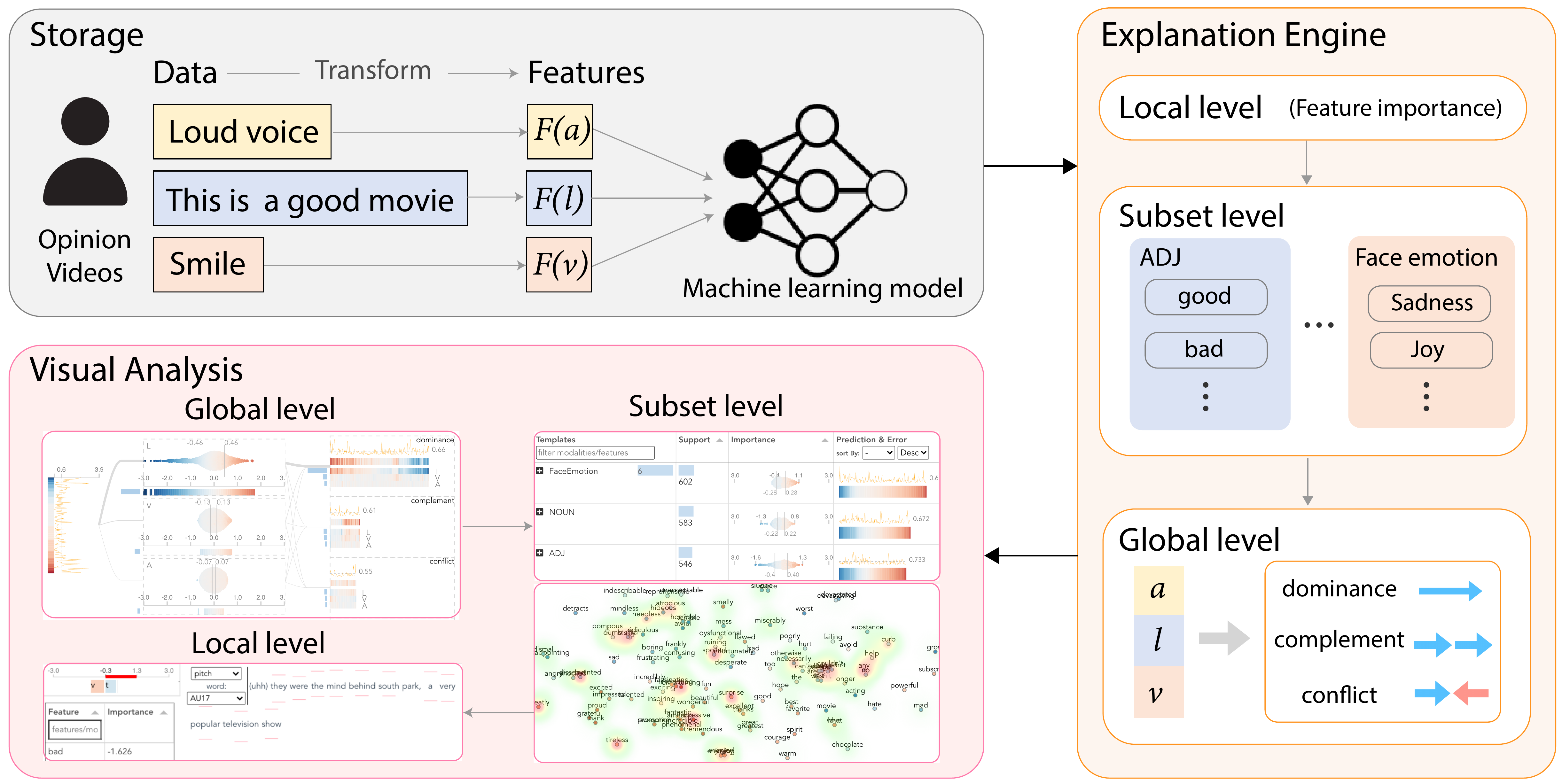}
\vspace{-5mm}
\caption{
% {\name} consists of three modules, a \textit{storage} module for the data and model, an \textit{explanation engine} for multi-level explanation generation, and a \textit{visual analysis} module for interactive exploration of model behavior.
{\name} consists of a \textit{storage} module, an \textit{explanation engine}, and a \textit{visual analysis} interface.}
\label{fig.system_framework}
\end{figure}

\autoref{fig.system_framework} shows the system architecture. 
\xbRevise{First, speakers' opinion videos are transformed into visual, acoustic, and language features. The \emph{storage} module saves users' model and data with processed features. Then, the \emph{explanation engine} inputs the features into the model and generates multi-level explanations of model behaviors based on the feature attribution methods (\eg, SHAP). The \emph{visual analysis} module enables interactive exploration of the explanations through five main views.}
% Our system contains three components, a \emph{storage} module, an \emph{explanation engine} module, and a \emph{visual analysis} module.
% The \emph{storage} module saves users' models and multimodal data with processed features.
% Given a model and the input features, the \emph{explanation engine} computes the feature importance using feature attribution methods (\eg, SHAP). Then, it generates multi-level explanations on model behavior.
% The \emph{visual analysis} module supports interactive exploration of the explanation results with five main views.

The {\vone} is the entry point of the whole interface, where the descriptive statistics about the model performance and dataset (\imp{Q1}) are shown.
Then, {\vtwo}, {\vthree}, {\vfour}, and {\vfive} provide multi-level model explanations from language, visual, and acoustic modalities (\imp{R4}).
The {\vtwo} presents a global summary of the influences of individual modalities and their interplay for the sentiment predictions (\imp{R2}).
The {\vthree} and {\vfour} complement each other for subset-level explanations (\imp{R3}).
Specifically, {\vthree} uses templates to summarize feature sets that frequently and significantly contribute to the model predictions. {\vfour} supports the multi-faceted exploration of instances that have features of interest, along with their prediction errors.
The {\vfive} summarizes instance-level prediction information (\eg, errors) (\imp{Q2}) and offers local explanations on the importance of each modality and its features (\imp{R4}). In addition, it adds the audio and vision features along the spoken words and provides the corresponding raw video clips with feature annotations for further exploration.

\subsection{Multi-level Explanations}
\label{subsec:explanation-generation}
% \muqiao{Can we change the subtitle since I think explanation generation is another research topic term}
To facilitate users with a comprehensive understanding of multimodal behavior, we propose methods to generate global and subset-level explanations (\imp{R2, R3}). 
They supplement the local explanations computed by feature attribution methods.

\subsubsection{Global Explanations}
Since the intra- and inter-modal interactions lie at the heart of multimodal sentiment analysis, they are essential for users to understand how the multimodal model utilizes the information from different modalities (\ie, language, audio, and vision) (\imp{R2}).
In our work, we characterize three typical types of interactions among modalities---dominance, complement, and conflict (details are in \autoref{subsec:modality-interaction}).

The \imp{dominance} suggests that the influence of one modality dominates the polarity (\ie, positive or negative) of a sentiment prediction.
The \imp{complement} indicates that two or all three modalities affect a model prediction in the same direction (\ie, positively or negatively). Conversely, the \imp{conflict} reveals that the influences of modalities differ from each other.
According to the definitions above, we formulate a set of rules to identify them (Algorithm \ref{alg:interaction-rule}). Specifically,
The influence of the interactions on the model output is based on the importance of each modality ($I_{l}, I_{a}, I_{v}$), which is the summation of the importance of all its features. 
% filter unimportant interactions with little influences, and 
Then, we extract and summarize the interactions ($L$) with strong influences for all the predictions. The thresholds for our rules are determined by 
maximizing the distances between the interaction types while minimizing the average influences of interactions that do not belong to dominance, complement, or conflict (\ie, others):

\begin{equation}
	\mathop{\arg\max}_{\{Th_{sig}, Th_{dom}, Th_{confl}\}}
	\frac{1}{|L|^2}\sum_{i}^{L}\sum_{j}^{L}{dist}(L_{i}, L_{j}) -  \bar{L}_{others}
\end{equation}

where 
% $\{Th_{sig}, Th_{dom}, Th_{confl}\}$ are the thresholds of Algorithm \ref{alg:interaction-rule} for important interactions, dominance, and conflict.
$L_i$~($i \in \{\imp{dominance, conflict, complement, others}\}$) is the interaction types output by Algorithm \ref{alg:interaction-rule} for all the instances, ${dist}$ is the Euclidean distance between the average influences of $L_i$ and $L_j$.

% \zhihua{
% 1. How to determine the distance between the group centers? Euclidean? What are group centers?
% 2. For the algorithm part, I cannot fully understand it. Maybe it lack of notations of $\exists i \forall j$. 
% 3. I think the label is a noun not an adj shown in algorithm part. 
% 4. Lack of explicit pointing what is the meaning of notations ${Th_\theta}$, $n$, $ \bar{I}_{others}$. 
% 5. Lack of describing how to normalize.
% 6. If there are some comments in the psudocode, it would be better. Line number should be added.}

\begin{algorithm}[htb] 
\caption{Rules for extracting important relationships of modalities.} 
\label{alg:interaction-rule} 
\begin{algorithmic}[1] %这个1 表示每一行都显示数字
\REQUIRE %算法的输入参数：Input
$\{I_{l}, I_{a}, I_{v}\}$;
$Th_{sig}, Th_{dom}, Th_{confl} (\in (0,1))$; \\
% /* \texttt{thresholds for important features, dominance, and conflict} */
% Ensemble of classifiers on former batches, $E_{n-1}$;
\ENSURE %算法的输出：Output
Label for the interaction types, $l$;
% \IF{$max(\{|I_{l}|, |I_{a}|, |I_{v}|\}) \leq Th_{sig}$}
% \STATE L = \imp{unimportant}; 
% \label{ code:unimportant}
\IF{$\forall i \in \{l,a,v\}, |I_{i}| > Th_{sig}$}
\STATE  /* \texttt{important interactions} */
\IF{$\exists i,j \in \{l,a,v\}, I_{i}\cdot\sum {I_j}>0, \frac{|I_{i}|}{\lVert I \rVert} \geq Th_{dom}$}
\STATE $l = \imp{dominance}$; 
\label{ code:dominance }
\ELSIF{$\exists i,j \in \{l,a,v\}, I_{i} \cdot I_{j}<0, \sum \frac{I_{i}}{\lVert I \rVert} \leq Th_{confl}$}
\STATE $l = \imp{conflict}$; 
\label{ code:conflict }
\ELSIF{$\exists i,j \in \{l,a,v\}, I_{i} \cdot I_{j}>0$}
\STATE $l = \imp{complement}$; 
\label{ code:complement }
\ELSE \STATE $l = \imp{others}$;
\label{ code:other }
\ENDIF
\ELSE
\STATE $l = \imp{others}$; 
\ENDIF
\end{algorithmic}
\end{algorithm}

% \IncMargin{1em}
% \begin{algorithm}
%   \SetKwData{Left}{left}\SetKwData{This}{this}\SetKwData{Up}{up}
%   \SetKwFunction{Union}{Union}\SetKwFunction{FindCompress}{FindCompress}
%   \SetKwInOut{Input}{input}\SetKwInOut{Output}{output}

%   \Input{A bitmap $Im$ of size $w\times l$}
%   \Output{A partition of the bitmap}
%       \lIf($\forall i \in \{l,a,v\}, |I_{i}| > Th_{sig}$){
%       }
%       \lElse{
%         $l = \imp{others}$
%       }
% \end{algorithm}\DecMargin{1em}

\subsubsection{Feature Templates}
\label{subsubsec: feature templates}

% \yong{Need to check it: explain how we group features.}

Compared with inspecting the impacts of individual features, exploring feature groups is more effective for analyzing complex model behaviors and data characteristics~\cite{krause2015supporting, kahng2017cti}.
It helps users develop a mental model about the model decisions (\imp{Q6}). For example, what types of words (\eg, adjectives) are considered important indicators for positive sentiment.
To ease the exploration of influences of high-dimensional features,
\xbRevise{we organize the model's input features introduced in \autoref{subsec: multimodal feature engineering} into several meaningful groups. Then, we summarize frequent and influential groups with compact templates (\xingbo{\autoref{fig:teaser}C}).}
% we organize low-level features into groups and summarize frequent and influential groups with templates (\xingbo{\autoref{fig:teaser}C}).
% \yong{We may need to further clarify what do we mean by saying ``templates'', e.g., refer to Fig. 1?}

\xingbo{To promote the understanding of model behaviors, we first identify several \emph{feature sets} based on the sentence structures for the language modality, emotion-related features for the acoustic modality, and facial expressions for the visual modality:}
% \zhihua{Whether we need to keep the style consistency within bulleted items?}
\begin{compactitem}
\item \textit{Language}: part of speech (POS)\footnote{\url{https://universaldependencies.org/docs/u/pos/}} (\eg, noun, adjective, verb);
\item \textit{Audio}: pitch, amplitude, glottal/voice quality, and phase;
\item \textit{Vision} (\ie, Face): face parts (\ie, brow, eye, nose, lip, and chin), head movement, and face emotions.
\end{compactitem}
% \yong{It is better to use a sentence to illustrate the definition/scope of language, audio, and vision.}
For language modality, POS features provide a compact summary of the structure of language use. They have been widely used as a probe for natural language models~\cite{ribeiro2020beyond, rogers2020primer, strobelt2017lstmvis}. %Since POS are not
The audio features are grouped according to a state-of-the-art speech processing framework, COVAREP~\cite{degottex2014covarep}, and speech applications~\cite{rubin2013content,wang2020voicecoach}. These sets generally relate to the emotions and tones of speech.
For face-related features, we divide them into the face parts, head movement, and face emotions. They are the representative components in the facial action coding system (FACS)~\cite{ekman1997face} for describing facial expressions. For the mapping between low-level multimodal features and the feature sets, please refer to \xingbo{the supplementary material}. 
% \zhihua{Where is supplementary material?}

After grouping the low-level features for each modality, we construct templates for both the frequent feature sets (\eg, ``\texttt{ADJ}'') and features (\eg, word ``good'') that have a strong influence on predictions (\imp{Q6}).
% (\autoref{fig.freq_set}).
% \yong{What are frequent feature sets? It is unclear.}
Specifically, we create itemsets of important features and feature sets for all predictions. Then, we build FP trees~\cite{han2004mining} to find frequent patterns within the itemsets. For example, if ``\texttt{PRON}'' and ``\texttt{PART}'' or  the word ``not'' constantly appear, they will be recorded in the templates (\xingbo{\autoref{fig:teaser}C}).

% \begin{figure}[htb]
% \centering 
% \includegraphics[width=0.5\textwidth]{figures/freq_set.pdf}
% \vspace{-5mm}
% \caption{
% Extraction of feature templates.
% } 
% \label{fig.freq_set}
% \vspace{-5mm}
% \end{figure}

\subsection{User Interface}
Based on the generated explanations, the user interface of {\name} facilitates
multi-level exploration of model behavior from the perspective of language, acoustic, and visual modalities (\imp{R4}). All the views are tightly integrated with interactions to ensure a smooth transition between different levels of explanations. They share the same color encoding scheme where dark red means strong positive sentiment and dark blue represents strong negative sentiment.
% at global, subset, and local levels.

% Since the intra- and inter-modal interactions lie at the heart of multimodal sentiment analysis, they are essential for users to understand how the multimodal model utilizes the information from different modalities (\ie, language, audio, and vision) (\imp{R2}).
% In our work, we compute the importance of individual modality and characterize three typical types of interactions among modalities---dominance, complement, and conflicts.
% Then, we visualize the results in a augmented tree-like layout to facilitate the global understanding of multimodal model behavior.
% The {\vone} presents the impact of intra- and inter-modal interactions on the model predictions (\imp{R2}).
% Specifically, it focuses on the individual modalities and the dominant, complementary, and conflicting relationships among them. Here, we will describe the generation of global explanations and the visual designs of {\vone}.

\begin{figure}[t]
\centering 
\includegraphics[width=0.47\textwidth]{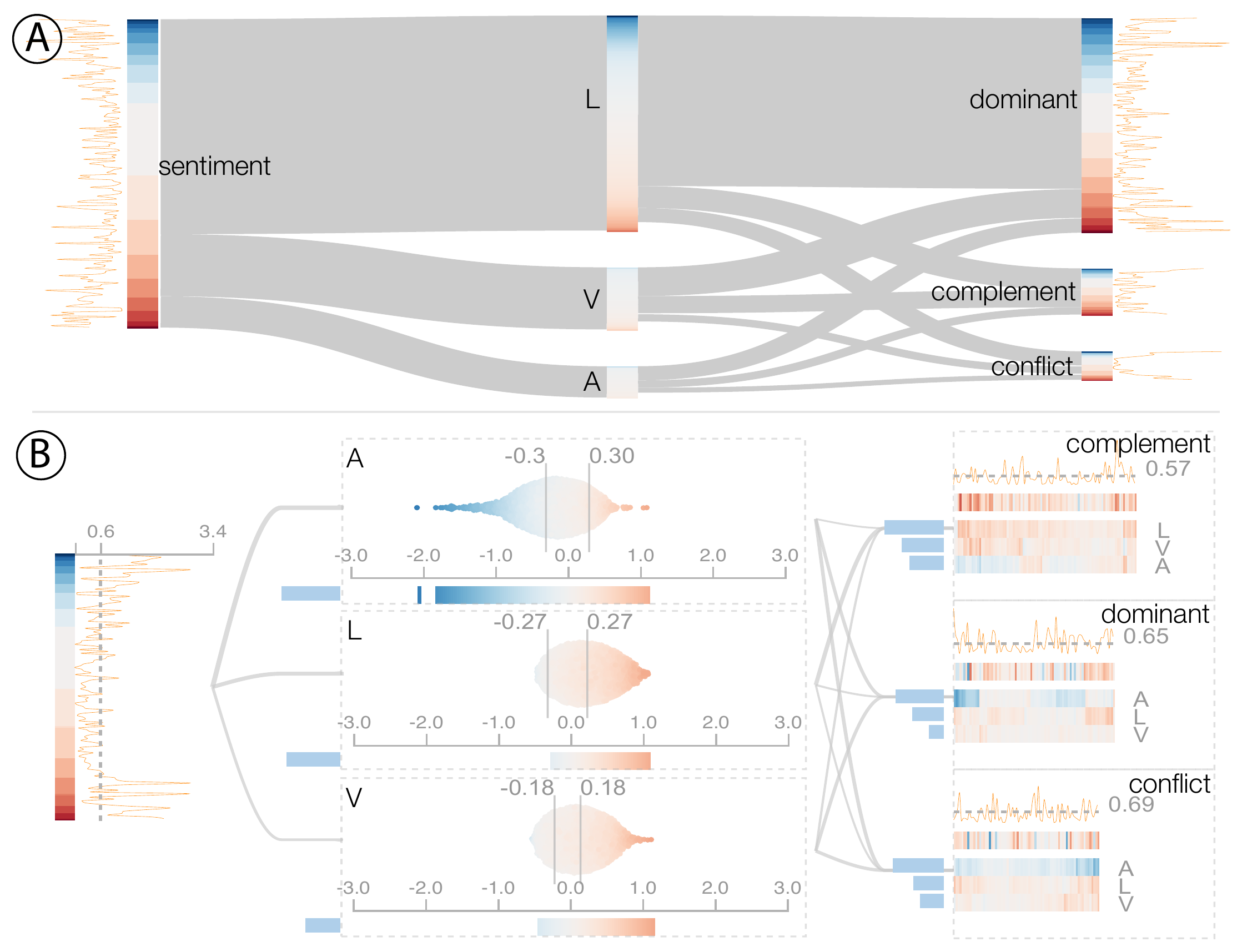}
\vspace{-4mm}
\caption{
Design choices for the {\vtwo}.
A: An augmented Sankey diagram.
B: Our current design of augmented tree-like layout. } 
\vspace{-5mm}
\label{fig.alternative_summary}
\end{figure}

\subsubsection{Summary View}
\label{subsubsec:summaryview}
% After producing the explanations for the selected data and model in the {\vone}, the 
The {\vtwo} presents an overview of the intra- and inter-modal interactions that are learned by the selected model in the {\vone} (\imp{R2}).
The influences of individual modalities and their interplay are visualized in a three-layer augmented tree-like layout (\autoref{fig.alternative_summary}B).

\textbf{Visual designs.}
In the parent node, a barcode chart and a line chart show the distributions of the ground truths and model prediction errors, respectively (\imp{Q1}). 
The vertical height of the barcode represents the total number of instances, and the color displays the sentiment. Meanwhile, the horizontal position of the line chart suggests the absolute error, \xingbo{and the mean error is represented as a dashed line}.

The second layer presents the importance of individual modalities in bee swarm plots (\imp{Q3}). They are arranged according to the influences of modalities in descending order. For each node in the layer, a blue bar is put to the left, whose horizontal length summarizes the total influences of the modality.
Besides, the dots in the bee swarm plot and their projections (\ie, the barcode below) demonstrate the distribution of the influences of that modality for all the instances. The color and horizontal position of the dots encode the importance values, while
the two gray lines indicate the magnitude of mean absolute importance. 
% the dot density indicates the frequency information. 
% Two gray lines indicate the mean absolute importance values.

% Based on the importance values of modalites, all the data instances are categorized into four types of interactions---dominance, complement, conflict, and others (\autoref{subsec:explanation-generation}).
The last layer summarizes the information about the four types of interactions~(\autoref{subsec:explanation-generation}), where the most influential one is shown at the top (\imp{Q4, Q5}).
For each interaction, 
the horizontal range of all its charts marks the number of instances in that group.
To better surface the patterns of how the combinations of modalities affect the model predictions, we put the data instances close to each other if all their three modalities share similar influence patterns. Specifically, the similarity is measured by the farthest distances among three modalities between the instances.
Then, 
a line chart and a barcode chart at the top summarize the error and prediction patterns, which are similar to the parent node.
In addition, three barcode charts are attached below to present the distribution of importance of all three modalities.
Their vertical orders show the total influences of the corresponding modalities, which are summed up by the blue bars to the left. The color of the bars inside the barcodes represents the importance values.

Besides, between two neighboring layers, links are drawn from the parent nodes to their child nodes. The width of a link is proportional to the importance of the child node to the model predictions.

\textbf{Design choices.}
\xingbo{We have considered an alternative design (\autoref{fig.alternative_summary}A) based on the Sankey diagram to reveal the intra- and inter-modal interactions and their importance to the predictions.
It consists of three parts, the ground truth information at the left, the influences of individual modality at the center, and the inter-modal interactions at the right. The width of a flow is proportional to the importance of the target node of the flow. 
The barcode chart of each node further displays the importance distribution.
In addition, the orange lines of the nodes show the error distribution to guide the exploration.
However, one expert commended that it would be necessary to demonstrate more detailed information on each node. For example, what modalities dominate the predictions, and what is the frequency? Therefore, we augment the nodes with graphs
and further convert the Sankey diagram into a compact tree-like layout, which leads to the current design (\autoref{fig.alternative_summary}B).
}

\subsubsection{Template View}
To facilitate the exploration of feature sets and their influences, the {\vthree}
(\jianben{\autoref{fig:teaser}C}) summarizes frequent and influential templates of multimodal features in a table (\imp{Q6}).

\textbf{Visual designs.}
The {\vthree} has four columns describing information about the template types, support, importance, and predictions and errors (\imp{R1, Q6}).
% Each row in the table corresponds to a specific 
The first column records the names of feature sets by default.
% By default, only the frequent and important feature sets are listed. 
If a feature set contains frequent and important features, a green bar will be placed to the right denoting the number of children for the feature set. Users can collapse the corresponding row for detail by clicking the {\faPlusSquare}.
The second table column displays the frequency for the templates. 
The distribution of the templates' importance and prediction information is visualized in the third and fourth columns. They share the same visual representations with the {\vtwo} (\autoref{subsubsec:summaryview}).
Users are enabled to sort the templates according to their support, importance, and errors.
In such a way, they can prioritize their efforts in diagnosing the complex model behavior.

\begin{figure}[t]
\centering 
\includegraphics[width=0.5\textwidth]{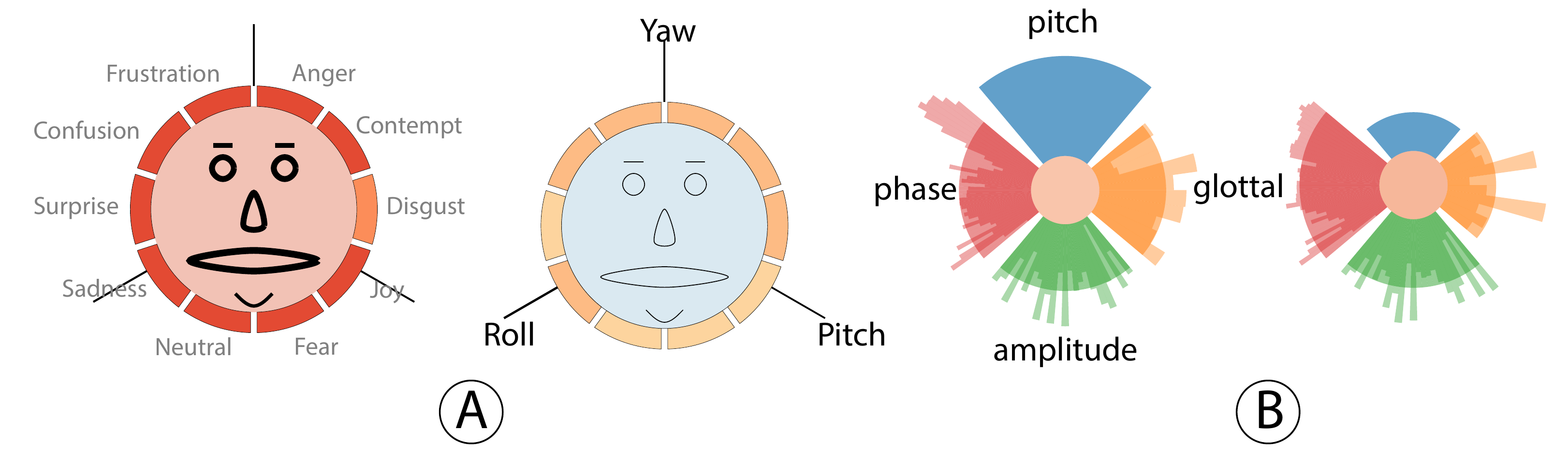}
\vspace{-6mm}
\caption{
The glyph designs in the {\vfour}.
A: Chernoff face glyph designs. The left one with darked colored rings and thick strokes of face parts indicates intense facial movement, while the right one suggests little facial movement.
% (left: intense facial movement, right: little facial movement).
B: Audio glyph designs. The left one with big blue sectors indicates high pitch, while the right one suggests low pitch.} 
\label{fig.glyph_design}
\vspace{-5mm}
\end{figure}

\subsubsection{Projection View}
\label{sec:embeddingview}
To further support the subset-level exploration of model behavior (\imp{R3, Q6}), the {\vfour} (\jianben{\autoref{fig:teaser}D}) connects the multimodal feature templates in the {\vthree} with the instances. 
It allows users to examine the detailed information (\eg, feature values, prediction errors) about features across the instances.
% reveals detailed information about the multimodal feature templates
% is designed and implemented. 
For example, after users select the ``\texttt{ADJ}'' template in the {\vthree}, 
% they may feel intrigued by ``fascinating'' and ``interesting''
% the instances of the template will be shown in the {\vfour}.
they may feel intrigued by what adjectives
% feature values (\ie, adjectives) 
associate with large errors
or with positive predictions. Then, they need to further inspect the individual instances.

% the {\vfour} will be updated to display the instances of corresponding templates. Users may find the instances with "cat", "dog" feature values with large errors on sentiment prediction. They may show interested in them and will further inspect individual instance information.

% motivation: projection view + template view (:frequent and important feature templates/subsets) => subset-level exploration of model behavior(R3: Q6)
% 比如: 得到了noun 这个template，我们可以通过embedding view 找到cat, dog (feature values)，并看它相应的instance的prediction以及error等信息。
% multi-faceted exploration
% 先介绍design思路，再mapping 特定的features
% 1. 
% for feature vectors (embeddings) == > refer to Sec. 3.2

\textbf{Visual designs.} 
To summarize the feature sets of a group of instances, we project the high-dimensional features onto a 2D plane using t-SNE~\cite{maaten2008visualizing}.
Thus, instances with similar features will be placed close to each other.
Given textual, acoustic, and visual features are heterogeneous, 
we design three different glyphs to encode the feature sets of the instances. Users can switch between views to see the feature distribution of each modality.
Moreover, to help diagnose the model behavior (\eg, errors), 
we add a heatmap as the background to display the distribution of prediction errors or template importance.

% To summarize the feature sets of a group of instances, we employ the t-SNE~\cite{maaten2008visualizing} algorithm to calculate the projection of selected instances to reveal the clusters. 
% In order to help users develop a high-level concept about model predictions. we further add a heatmap as the background to display the distribution of error or importance values. 
% Since the features of the three modalities are heterogeneous, we design three different glyphs to encode different feature sets of individual instances. Users can switch between different views to see each modality feature distribution. 
% view background (heatmap): error \& importance (shap values)
% Three different glyphs are implemented to help users further analyze. 
\begin{compactitem}
\item \textit{Language}: 
since words already carry semantic meanings, we use them to represent the textual features. In addition, we add a circle for each word, whose color encodes the sentiment prediction.
% In this glyph, the user may want to check the sentiment prediction for this instance and the corresponding text for it. The color of the circle encodes the sentiment prediction. The position is computed based on its GloVe embedding. 
%The position is determined by t-SNE algorithm applied for its GloVe embedding. For details of its embedding, please refer to \autoref{subsec: multimodal feature engineering}. %color: sentiment prediction) glove embedding tSNE
\item \textit{Vision}: our glyph designs for facial features (\autoref{fig.glyph_design}A) are inspired by Chernoff face~\cite{chernoff1973use}, which is popular for displaying facial expressions.
However, the original Chernoff face cannot reflect information such as head movement.
Therefore, we add three sticks around the face to indicate the head movement in the yaw, pitch, and roll axis, respectively. The outer ring encodes the whole face information (\eg, emotion in our case), where the dark color suggests large feature values.
Moreover, the stroke width of face parts (\eg, nose) and sticks mean movement intensity.
The sentiment prediction is revealed by the face's background color.
% \xingbo{For example, compared with the right face in A, the left one with thicker face parts and darker outer rings suggest more intensive facial muscle movement.}

% We use Chernoff face~\cite{chernoff1973use} to encode the facial expressions, demonstrated in ~\autoref{fig.glyph_design}(a). The thickness of each face part represented its movement. However, Chernoff face cannot reflect the head movement information. Three lines are further added in the glyphs and the length of those lines encodes the head movement in yaw, pitch, and roll axis respectively. The colored outer ring encodes the whole facial information. %Eyebrow are called face parts, which can be captured by Chernoff face.  % the  (glyphs: chernoff face (reference:) ==> facial expressions, three lines === head movement, outer rings === other face information (\ie, emotions) [However, chernoff face cannot reflect the head movement information. X, Y, Z, Colored outer ring encodes the whole facial information. Eyebrow are called face parts, which can be captured by chernoff face. ]

\item \textit{Audio}: to help understand acoustic features, we group them into higher-level classes (\autoref{subsubsec: feature templates}). As shown in \jianben{\autoref{fig.glyph_design}B}, each colored sector represents the features of a class, where the radius relates to feature values.
The sectors at the front summarized the average values of normalized features, while the small ones at the back display detailed feature values of the classes.
Additionally, the inner circle color shows the sentiment prediction.
% \xingbo{write here}
% For audio feature sets, it can be further grouped into four classes. In the glyph shown in~\autoref{fig.glyph_design}(b), for each class, the outer ring encodes the information (average normalized features) of feature sets and the inner ring encodes the information of individual normalized features of feature sets. Similar to text glyph, the inner circle encodes the sentiment prediction for that instance. 
\end{compactitem}

%Some interactions are added within this view to enhance the usability. %When users hover the glyph, correspondings information will be shown. 

% Interactions: Hover, Switch between different views to see each modality feature distribution.
% === lasso === 不用讲

\subsubsection{Instance View}
The {\vfive} (\jianben{\autoref{fig:teaser}E}) provides local explanations by visualizing the important multimodal features and the context (\ie, transcripts and videos) of individual instances (\imp{Q7}).

% is designed to present detailed feature information at instance level 
% and provide necessary local explanations for selected instance or instance groups (\imp{Q7}). 

\textbf{Visual designs.}
The left column presents a visual summary of the influences of modalities on the model predictions, as well as the prediction errors. 
Users can sort the instances according to different criteria (\eg, error) at the header and prioritize their efforts in instance-level exploration.
In each row, the horizontal axes demonstrate the sentiment range, where the prediction and ground truth are marked. Between the two values, the thick red line suggests the error.
Below the prediction mark, three colored rectangles represent the aggregated feature importance values of the three modalities. The length and color of each rectangle encode the magnitude and sign of importance. For example, the modality with negative influences on the prediction will be encoded by a blue rectangle and placed at the right. In addition, 
the feature table below allows users to sort and search for the importance values of features or modalities.

To promote a comprehensive understanding of the context of individual instances, 
the right column highlights the important features of the instances.
Unlike intuitive texts, the acoustic and visual features are harder to recognize. Thus, we align them with the spoken words and draw the most important ones using \xingbo{orange lines}.
The lines above the words correspond to acoustic features, while the lines below represent the visual features. The vertical offset of the lines denotes the feature values, and hence the fluctuations indicate the feature variation. 
In addition, the backgrounds of texts or feature lines reflect the importance of multimodal features at a word level.

% Here every column exhibits the information of one single instance and the users can scroll up and down to browse all instances previously selected in {\vthree} and {\vfour}. In every single column, the upper left axis ranging from -3.0 to 3.0 gives an overview of the ground truth and prediction value as well as the red-line denoted error(distance) between these two values. \jianben{how to differentiate gt and prediction value?(the color? how to describe?} The three colored rectangles under the axis respectively represent the aggregated feature importance values of the three modalities where the length and color of each rectangle encodes the corresponding importance value size and its sentiment affect. The below feature table further lists the importance value of every feature where users can sort them according to the value or directly search the feature or modality in the search box to quickly find the interested feature or feature group. The right side of the column is the text of the instance where the background color of each word encodes its importance value. Unlike the readily understood text, the vision and audio features are relatively hard to be recognized by human and thus we align the vision and audio features with the words and visually represent them on the top and the bottom of the text using the pink line. The vertical offset of the line denotes the feature value and hence the variation trends of single vision or audio feature can be easily perceived by users. 

The {\vfive} also provides video context for instance-level exploration.
% supports the video playback. 
When users click on the rows of the table, the corresponding video clips will pop up and play.
% so that users can closely examine the raw data for better understanding and diagnosis. 
To make the visual features more intuitive, the top-ranked facial features (sorted according to importance value) are highlighted with bounding boxes that cover the corresponding parts of the face.
Users can further find the detailed facial action units and their concrete meanings by hovering on the boxes.

\subsubsection{User Interactions}
The {\name} provides a rich set of interactions, which help unify the different views and facilitate multi-level and multi-faceted exploration with details on demand. 

\textbf{Brushing.}
Users can brush the barcodes in the last layer of {\vtwo} to emit a query on the specific data instances of an interaction type. Then, the {\vthree} and {\vfive} will show the related templates and local explanations, respectively.

\textbf{Clicking.}
Many interactions in the system can be triggered and undone by clicking.
For example, clicking the table rows in the {\vthree} will filter the irrelevant instances in the {\vfour} and {\vfive}.
Users can switch between feature projections of different modalities by clicking the radio buttons in the {\vfour}.
% Users can click in the first column of the {\vthree} to collapse the feature sets of interest for detail. 
When clicking the table rows in the {\vfive}, the corresponding instances in the {\vfour} will be shown, and its video clips will pop up and play.
In addition, users can click on the header of the {\vthree} and {\vfour} to undo the previous selections.

\textbf{Lasso and semantic zooming.}
To facilitate scalable exploration, users can use lasso or semantic zoom to focus on specific instances of interest in the {\vfour}. Then, the detailed information will be displayed in the {\vfive}. 
% After users lasso a group of instances in the {\vfour}, the detailed information will be displayed in the {\vfive}.

\textbf{Searching, sorting, and filtering.}
To narrow down the exploration space, users can sort and search for the instances or features in the table of {\vthree} and {\vfive}.
By adjusting the sliders in the {\vfour}, users can filter the instances according to the sentiment predictions and the feature importance of specific modalities.
\section{Evaluation}
In this section, we demonstrate how {\name} helps users understand and diagnose multimodal models for sentiment analysis through two case studies and interviews with \xingbo{three} domain experts (\imp{E1}, \imp{E2}, and \imp{E3}) using the CMU-MOSEI dataset.
\imp{E1} and \imp{E2} are 
NLP researchers who have multiple top research publications on multimodal language analysis (\eg, emotion recognition).
% whose expertise is in multimodal language analysis (\eg, emotion recognition).
% Both of them have published papers in top NLP conferences.
\imp{E3} is a senior software engineer who has five years' experience in developing affective computing applications.
\xingbo{The two cases are discovered by \imp{E1} and \imp{E2} during the system exploration in the interviews.
% The two cases are discovered by the experts during their exploration of using {\name} to analyze the multimodal sentiment analysis models in the expert interview.
The detailed feedback from all the experts is also collected and summarized.}

\subsection{Case One: Multimodal Transformer}
\label{subsec: case one}
% \yong{Yong's task}
% \yong{Start here.}

In the first case, the expert \imp{E1} explored and diagnosed a state-of-the-art model, Multimodal Transformer (MulT)~\cite{tsai2019multimodal}, for sentiment analysis using the CMU-MOSEI dataset (\autoref{subsec: dataset}).
\xingbo{MulT fuses multimodal inputs with cross-modal transformers for all pairs of modalities, which learn the mappings between the source modality and target modality (\eg, vision $\rightarrow$ text). Then, the results are passed to sequence models (\ie, self-attention transformers) for final predictions.}
All the multimodal features of the input data are aligned at the word level based on the word timestamps.
Following the settings of previous work~\cite{tsai2019multimodal}, we trained, validated, and evaluated MulT with the same data splits (training: 16,265, validation: 1,869, and testing: 4,643).
The details about the MulT are included in the supplementary material.

During the exploration, \imp{E1} observed that the language modality often dominates the predictions, and the model cannot
% handle well with 
handle
the negations in sentiment analysis very well. 
He further investigated the dominance of visual modality, where ``\texttt{Joy}'' and ``\texttt{Sadness}'' (two facial emotions) frequently co-occur. 
It was thought to be caused by the intense facial muscle movement, which was also captured by the model.

\subsubsection{Dominance of Language Modality}
\label{sec-dominance-lang}
\textbf{Global summary (\imp{R1, R2})} 
~After selecting the MulT and valid set in the {\vone}, 
\imp{E1} felt interested in how individual modalities and their interplay contribute to the model predictions.
By looking at the second layer of the {\vtwo} (\autoref{fig:teaser}B), \imp{E1} found that the language modality (indicated by the letter ``L'') has the largest influence among the three modalities since it has the longest bar to the left and widest range of dots in the bee swarm plot.
On the contrary, the acoustic modality (indicated by the letter ``A''), which ranks at the bottom, has the least influence.
Then, \imp{E1} examined the last layer, where the \imp{dominance} group with the widest barcode charts is shown at the top.
Within the group, he discovered that
the longest bars attach to the language modality, and
the color of the prediction barcode aligns well with that of the language barcode.
Thus, \imp{E1} concluded that the language also plays a leading role in the \imp{dominance} relationship.
Furthermore, he noticed that
% a group of blue bars concentrate at the end of the language barcode with relatively large errors (above the dashed line).
there are a group of dense blue bars appearing at the end of the language barcode, where the errors are relatively large (as indicated by the yellow curve above the dashed line).
He wondered what features or their combinations cause the high errors. Therefore, we brush the corresponding area of the blue bars.

\textbf{Subset exploration (\imp{R1, R3, R4})}
~The {\vthree} (\jianben{\autoref{fig:teaser}C}) lists all the frequent and important feature templates for the brushed instances in the {\vtwo}.
By sorting them in descending order of error, \imp{E1} found that the ``\texttt{PRON + PART}'' appears at the top with one child feature. Then, he collapsed the row and found that 21 instances contain the word ``not'', where it negatively influences the predictions (blue dots in the bee swarm plot in the ``importance'' column).
Next, he clicked ``not'' to see the details about this feature in the {\vfour}.
Zooming in on the word ``not'', several similar negative words (\eg, ``isn't'', ``wouldn't'') were observed (\xingbo{\autoref{fig:teaser}D}). They were all located in a red area, indicating large errors.
\imp{E1} speculated that the model cannot deal well with negations.
Subsequently, he lassoed these words to closely examine the corresponding instances in the {\vfive}.

\begin{figure}[t]
\vspace{-2mm}
\centering 
\includegraphics[width=0.4\textwidth]{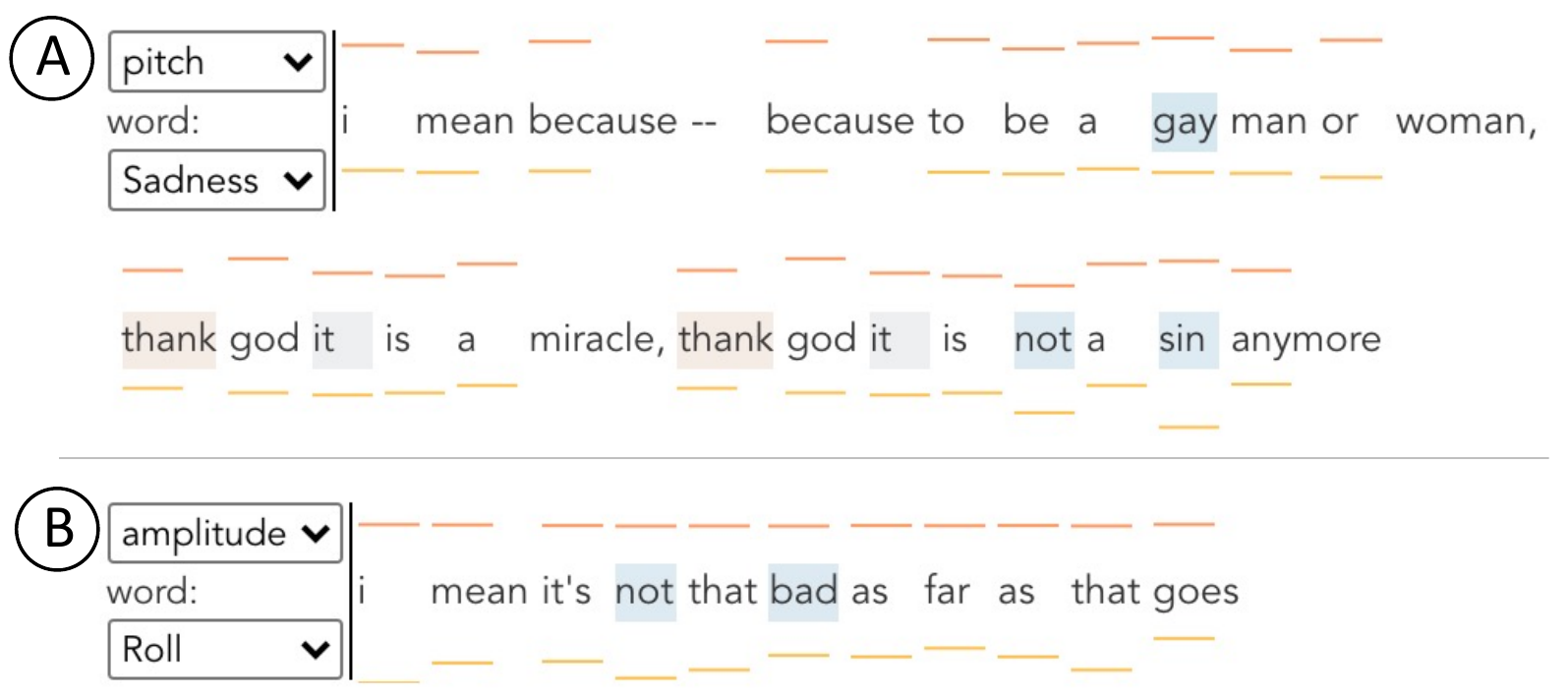}
\vspace{-3mm}
\caption{Examples of double negations.
``not...sin'' (in A) and ``not..bad'' (in B) are considered as indicators for negative sentiment by the model. However, these phrases reflect sentiments that are slightly positive.
} 
\label{fig.caseOneText}
\vspace{-5mm}
\end{figure}

\textbf{Instance exploration (\imp{R1, R3, R4})}
% To further evaluate the negation handling ability of the model,
~To further evaluate how the model handles negations,
% the ability of the model in handling negations,
\imp{E1} started with the instances with large errors in the table (\xingbo{\autoref{fig:teaser}E}).
When exploring the top-listed examples, \imp{E1}
observed that negations always have significant negative influences on the predictions, and the model fails to interpret the true sentiment.
For example,
% as shown in \xingbo{\autoref{fig.CaseOneText}-4}, 
\imp{E1} found a case where the language modality dominates the negative sentiment prediction, and the word ``not'' is highlighted in blue (\xingbo{\autoref{fig:teaser}E}). 
However, the true sentiment of this sentence is positive, where the starting phrase ``I really like'' demonstrates the positive attitude. 
However, the model fails to extract the keywords and relies on the negation (\ie, ``not'') to predict the negative sentiment.
Moreover, \imp{E1} noticed that when double negations appear in a sentence 
% (\eg, ``not...stagnant'', ``not...sin'' in 
(\xingbo{\autoref{fig.caseOneText}}), 
the model tends to treat them separately and regards both of them as indicators for negative sentiment. However, in fact, these double negations reflect sentiments that are slightly positive.

% \begin{figure*}[!t]
%       \setlength{\belowcaptionskip}{-5pt}
%   \centering
%   \vspace{-3mm}
%   \includegraphics[width=1.92\columnwidth]{figures/caseOneVision.pdf}
%   \vspace{-3mm}
%   \caption{``\texttt{Joy + Sadness}" co-occurrence patterns. A: ``\texttt{Joy + Sadness}" is a frequent and important feature template in the table. B: The raw video information and corresponding glyphs of three representative instances of the ``\texttt{Joy + Sadness}" template.}
%   \label{fig.CaseOneVision}
%   \vspace{-1em}
% \end{figure*}

\subsubsection{Dominance of Visual Modality}
~\textbf{Global summary (\imp{R1, R2})}
~\imp{E1} referred back to the ``\imp{dominance}'' group in the {\vtwo},
where a collection of red bars from the prediction barcode conform with the ones from the visual modality (highlighted red in \jianben{\autoref{fig:teaser}B}).
The visual modality dominates the predictions, and the error line chart above suggests a low error rate in contrast with the previous case in Sect.~\ref{sec-dominance-lang}.
Motivated by this observation, \imp{E1} brushed the red bars to investigate the patterns in the visual features.

\begin{figure}[t]
\vspace{-2mm}
\centering 
\includegraphics[width=0.45\textwidth]{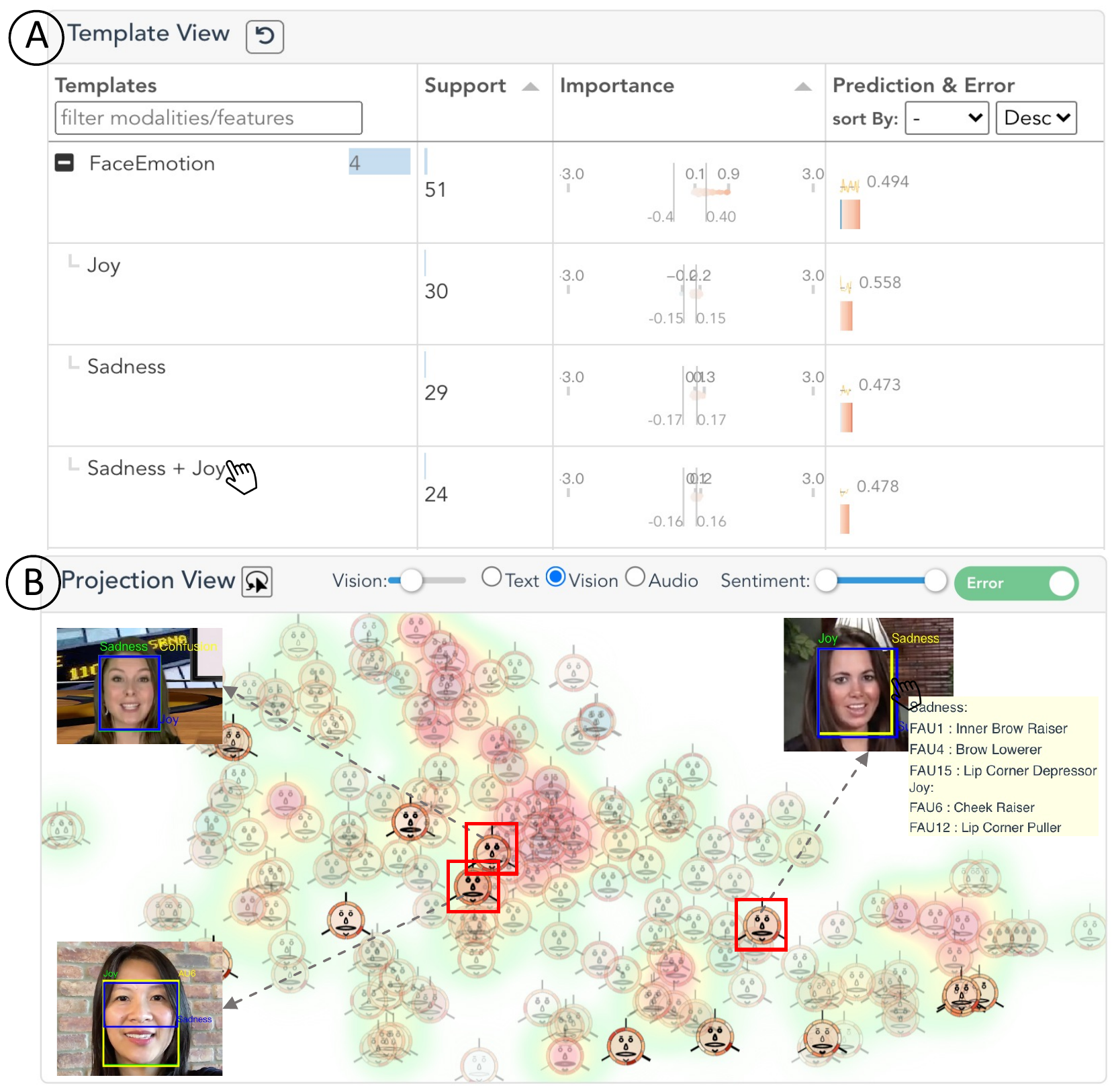}
\vspace{-3mm}
  \caption{``\texttt{Joy + Sadness}" co-occurrence patterns. A: ``\texttt{Joy + Sadness}" is a frequent and important feature template in the table. B: The raw video information and corresponding glyphs of three representative instances of the ``\texttt{Joy + Sadness}" template.}
  \label{fig.CaseOneVision}
  \vspace{-1.5em}
\end{figure}

\textbf{Subset exploration (\imp{R1, R3, R4})}
~In the {\vthree}, ``\texttt{Face Emotion}'' has the largest support (\xingbo{\autoref{fig.CaseOneVision}A}). After unfolding the row, \imp{E1} found that ``\texttt{Joy + Sadness}'' is a frequent and important combination.
This intrigued him to find out how a contrary emotion pair co-occurs.
After clicking the template, the corresponding glyphs are highlighted in the {\vfour} (\xingbo{\autoref{fig.CaseOneVision}B}). Most of them are found outside the red area, which verifies that the instances with``\texttt{Joy + Sadness}'' 
% associate with 
often have
small prediction errors.
% Then, 
He decided to inspect these instances.

\textbf{Instance exploration (\imp{R1, R3, R4})}
~Through browsing the instances and their videos in the {\vfive}, ``\texttt{Joy}'' and ``\texttt{Sadness}'' are often considered important visual features with positive influences.
Additionally, \imp{E1} found their co-occurrences may be due to the presence of intense and rich facial expressions in the videos. 
These expressions generally involve the movement of the related facial action units in ``\texttt{Joy}'' and ``\texttt{Sadness}''.
For example, 
% as shown in (\xingbo{\autoref{fig.CaseOneVision}-4}), 
% ``\texttt{Joy}'' and ``\texttt{Sadness}'' are the top important visual features in the table (\xingbo{add fig., maybe will remove feature table in the figure to make space}).
after \imp{E1} clicked on the instances, he noticed
all the face parts (\ie, nose, eyes, brows, mouth, and chin) of the corresponding glyphs (\xingbo{\autoref{fig.CaseOneVision}B})
in the {\vfour} has thick strokes, which suggests intense movements.
When he watched the original videos, the bounding boxes of ``\texttt{Joy}'' and ``\texttt{Sadness}'' always popped up as important visual features. Hovering on the boxes and examining the facial expressions and their explanations, \imp{E1} concluded that the extreme facial expressions triggered the movement of the action units in ``\texttt{Joy}'' and ``\texttt{Sadness}'', and the model seemed to capture these important visual facial expressions.

%%%%%%%%%%%% draw conclusions %%%%%%%%%%%%%%%%%%%%%%
\xbRevise{\paragraph{Conclusions.}During exploration,
% exploring MulT with \name{}, 
\imp{E1} discovered that MulT
% has limitations of handling 
cannot handle
double negations very well, though it is a state-of-the-art model. He commented augmenting double negation examples or preprocessing them into positive forms can further improve the performance.}

\subsection{Case Two: EF-LSTM}
\label{subsec: case two}

% \yong{Yong's task}

In this case, the expert \imp{E2} explored the popular RNN-based model, EF-LSTM~\cite{hochreiter1997long}, for multimodal sentiment analysis using the CMU-MOSEI dataset.
The dataset setup and feature processing are the same as Case One (\autoref{subsec: case one}).
EF-LSTM concatenates textual, acoustic, and visual features at each word. Then, it uses an LSTM model to derive the input representations for the predictions. The details of the model are provided in the supplementary material.
% \yong{Make sure we will add it in the supplementary material. If no, just remove the sentence.}

Through interactive explorations with {\systemname}, 
% \imp{E2} found that EF-LSTM does not learn sentiment in text.
% To his surprise, the acoustic modality has the largest influences among three modalities. 
% Meanwhile voice pitch always plays an negative role on the sentiment predictions.
\imp{E2} was surprised to find that EF-LSTM does not learn sentiment in text.
Also, he noticed that the acoustic modality has the largest influence on the sentiment prediction results among the three modalities, and the voice pitch always plays a negative role in the sentiment predictions.

\subsubsection{No Meaningful Information Learned in Text}

\textbf{Global summary (\imp{R2})}
~After selecting the valid set and EF-LSTM, 
\imp{E2} started with the {\vtwo} to gain an overview of the impacts of the modalities (\xingbo{\autoref{fig.alternative_summary}B}).
By comparing the range of dots in the three bee swarm plots, he was surprised to find that acoustic modality is the most influential modality, then comes the language modality.
In addition, the language modality always exhibits a positive impact on the sentiment.
These findings are quite counter-intuitive.
Thus,
\imp{E2} first explored text-related interactions by tracking the thickest links from the language modality to the third layer. He noticed that ``\imp{complement}'' group shows at the top, and the text plays a leading role within the group. Then, he brushed the whole group to see textual feature patterns.

\textbf{Subset exploration (\imp{R3, R4})}
~The strange thing is that no textual templates and text glyphs were spotted in the {\vthree} and the {\vfour}, respectively.
\imp{E2} suspected that the model does not learn any important language features (\ie, words) for sentiment analysis.
Then, he referred to the {\vfive} to validate his doubt.

\qbox{It's run by a fantastic team of professors; they are always available for you. \\
(Umm) this movie was excellent.}

\textbf{Instance exploration (\imp{R1, R3, R4})}
~When exploring the instances in the {\vfive}, 
% \imp{E2} discovered that no words are highlighted in the \textit{Instance Detail}. 
\imp{E2} found that the model fails to recognize potentially-important words for sentiment analysis, such as ``fantastic'' (in line \#1), ``excellent'' (in line \#2).
None of them is highlighted with colors in the \textit{Instance Detail}.
\imp{E2} also noticed every word of the sentences in the feature table has evenly low positive importance scores (less than 0.1). 
This explains why the language modality always has positive influences and further proves that the model does not capture the sentiment in text. 

% \textbf{Is language learnt or not?} 
% First we look at the Summary View where the beeswarm plots of the three modalities imply that text always positively effect the sentiment prediction while audio play negative role for most of time. This discovery seems unreasonable (because ...) and thus we firstly choose the complementary box where text is the most significant channel to inspect its inner mechanism. The strange thing is the corresponding templates show in templates view exhibit no related text templates and the embedding view cannot filter any group of texts even we tune the filtering threshold to the minimum value. From these cues, we suspect that the model didn't learn anything important from the text. Then we go into feature states view to inspect individual instances to validate our doubt. Through browsing the instances, we discover that there is no color highlighting area in the script, which means the model fails to  recognize the potential important words for sentiment analysis like "excellent", "great", "awful" and "atrocious". Moreover, from the detailed feature importance table of each instance, we can see that every single word in the text has very low(less than 0.1) but relative evenly-distributed feature values which further proves that the model does not have the capacity to understand words. Therefore, for some long scripts, the simply accumulated importance values will exaggerate and misinterprete the text importance due to the lack of text understanding.

\subsubsection{Negative Influences of Voice Pitch}

\begin{figure}[!t]
\vspace{-2mm}
\centering 
\includegraphics[width=0.45\textwidth]{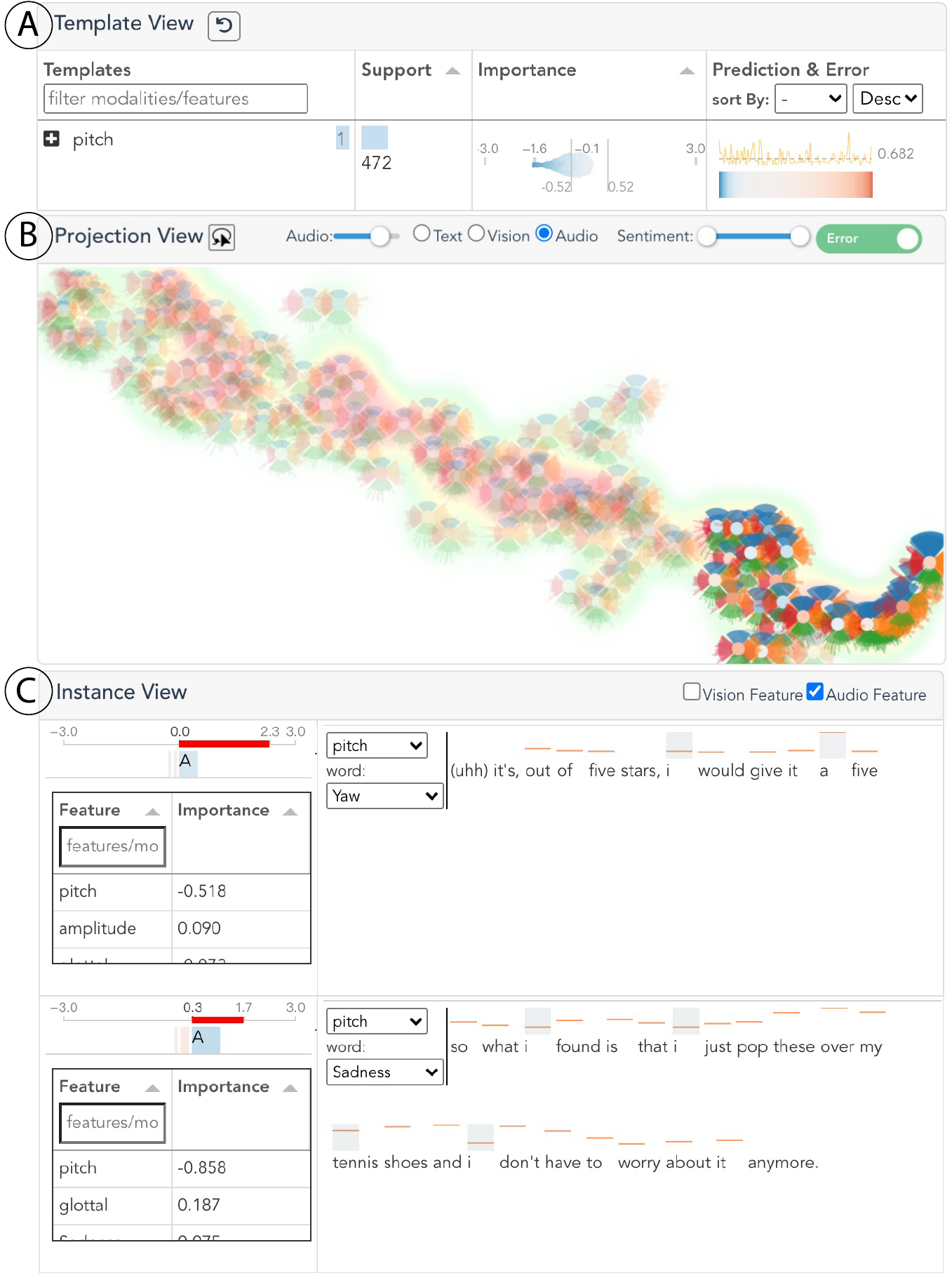}
\vspace{-3mm}
  \caption{Negative influences of voice pitch. A: ``\texttt{pitch}" is the most frequent acoustic template, and it always has a negative impact (as indicated by the dots in the bee swarm plot). B: The selected group of instances with large pitch values (as indicated by the large radius of the blue sectors). C: Two high-error cases where the model captures the turning points of the pitch but wrongly associates pitch with negative influences.}
\label{fig.CaseTwoAudio}
\vspace{-1.5em}
\end{figure}

\textbf{Global summary (\imp{R1, R2})}
~\imp{E2} paid attention to the most influential modality (i.e., the acoustic modality) in the {\vtwo} (\autoref{fig.alternative_summary}B), where a negatively-skewed distribution of dots was shown.
In addition, he noticed that within the``conflict'' group, the acoustic modality plays a negative role (blue bars) throughout the time.
Thus, \imp{E2} brushed this group to investigate the negative influence of acoustic features.

\textbf{Subset exploration (\imp{R1, R3, R4})}
~\imp{E2} found the ``\texttt{pitch}'' is the most frequent acoustic template in the {\vthree} (\xingbo{\autoref{fig.CaseTwoAudio}A}).
Moreover, \imp{E2} noticed that pitch always has a negative impact given the negatively-skewed distribution of dots in the third column.
After clicking the row, he switched to the {\vfour} to see the pitch value distribution (\xingbo{\autoref{fig.CaseTwoAudio}B}).
He discovered that the acoustic glyphs are spread along a left-slanting line, where the radius of the blue sectors (\ie, pitch values) generally increases from left to right. 
Then, he selected a group of instances with the large pitch at the right corner for further inspection.

\textbf{Instance exploration (\imp{R1, R3, R4})}
~By browsing the instances and videos in the \textit{Instance Summary} (\xingbo{\autoref{fig.CaseTwoAudio}C}),
\imp{E2} observed that pitch is always the top important acoustic feature and is associated with negative influences.
Although some important pitch variation signals in the videos are captured by the model, 
% he generally thought 
he believed that
the model is not reliable since it always regards the pitch as a strong negative sentiment indicator and he found many counterexamples.
To name a few, in two cases (\autoref{fig.CaseTwoAudio}C),
% \yong{What are the two cases here? It is a bit ambiguous.}
he found pitch ranks the first with negative importance in the feature table. 
And he noticed that some backgrounds of the orange lines (\ie, pitch values) are colored light blue (\ie, negative). By examining the offsets of all the orange lines, he thought the highlighted ones seem to be the turning points of pitch values. He speculated that the model captures the important signals in audio.
He further checked the original video and verified the observations.
However, the speakers sound high-spirited, and the pitch should reflect positive sentiment.
% \yong{pls check my side comments.}

%%%%%%%%%%%% draw conclusions %%%%%%%%%%%%%%%%%%%%%%
\xbRevise{\paragraph{Conclusions.} Through the case study, \imp{E2} found that EF-LSTM seems not able to capture the sentiment in text. He reasoned that the simple early feature fusion may lead to textual information loss. He 
% suggested 
speculated that some
more advanced model designs (\eg, transformer) can be incorporated into the model to facilitate text understanding. 
Given the negative impacts of voice pitch, \imp{E2} thought that removing the pitch feature may increase the model accuracy.}

\subsection{Expert Interviews}

We collected the feedback from the \textit{one-on-one} interviews with the aforementioned three domain experts (\imp{E1, E2, E3}).
% \imp{E1} and \imp{E2} are 
% NLP researchers who have multiple top research publications on multimodal language analysis (\eg, emotion recognition).
% % whose expertise is in multimodal language analysis (\eg, emotion recognition).
% % Both of them have published papers in top NLP conferences.
% \imp{E3} is a senior software engineer who has five years' experience in developing affective computing applications.
None of them have tried the system before the interviews.
We 
% spent the first 20 minutes introducing 
first introduced
the background and system designs.
Then we asked the experts to use {\name} to diagnose two state-of-the-art models (\ie, multimodal transformer and EF-LSTM) on the CMU-MOSEI dataset.
After a 50-minute exploration, we collected their feedback about the system workflow, system designs, application scenarios, and improvement suggestions.
% \yong{Do we do a one-to-one interview or a one-to-many interview? It is better to explicitly clarify it.}

\textbf{System workflow.}
All the experts
% agreed on the reasonableness and 
confirmed the
effectiveness of the system workflow of {\name} in providing explanations for multimodal sentiment analysis models.
They mentioned that they usually rely on performance metrics or instance-level feature importance measures for model evaluation, which does not provide many details and is unable to support an in-depth analysis.  
Our system supplements them with global- and subset-level explanations, which facilitates a comprehensive and systematic understanding of model behaviors.
\imp{E1} and \imp{E3} praised that the interaction summaries (\ie, \emph{dominance}, \emph{complement}, and \emph{conflict}) are impressive and very useful for revealing both the model behaviors and the multimodal data characteristics.
\xbRevise{\imp{E3} mentioned if he finds some modalities are influential in predicting sentiment using {\name}, he can consider reducing the number of modalities without losing much performance when deploying the model to low-end devices.}
\imp{E1} added that the feature templates help generalize the model error patterns.
\imp{E2} summarized that the system assisted him in discovering interesting insights into the models. For example, he was surprised that EF-LSTM 
% does not capture the sentiment in text.
seems to not capture any sentiment information from the text.

% \begin{compactitem}
% \item They mainly focus on instance-based explanations, while our system greatly supplement it with global and subset level understanding, which can help them generalize their findings about the model behavior.
% \begin{compactitem}
% \item The three interaction types (\ie, conflict, complement, and dominance) are interesting and reasonable for diagnosing multimodal models
% \item feature templates are intuitive and 
% \end{compactitem}

% \item The system help them find some unexpected things (\imp{E2} for EF-LSTM)
% \end{compactitem}

\textbf{Visual designs and interactions.}
Overall, the experts confirmed that the visualizations are useful and still easy to understand, and interactions are smooth.
% \yong{It has a conflict with our discussion on learning curve.}
The {\vtwo} is most favored by the experts for a quick overview of the learned intra- and inter-modal interactions.
The designs of {\vfour} are also appreciated by the experts.
\imp{E3} really liked the heatmap for showing the error and feature importance patterns.
\imp{E1} thought the face glyphs are very intuitive, and the interactions such as lasso and zoom are really helpful for the exploration of a large amount of data.
Moreover, he valued the video playback and the realtime highlighting of face parts for raw video browsing.
Nevertheless, 
\imp{E1} and \imp{E2} said that the {\vfive} is a little complex,
% and overwhelming, 
visualizing lots of information.
Additionally, the experts responded that it took them a while (\xingbo{about 20 minutes}) to fully grasp all the components and functions in the system. 
% \begin{compactitem}
% \item summary view: most liked
% \item projection view: heatmap intutive and useful for exploring error patterns, lasso, zoom are interesting; glyph designs are intuitive (face glyphs are praised by \imp{E1})
% \item video annotations are useful
% \end{compactitem}

% \xbRevise{\textbf{Application scenarios}. All the experts showed great interest in using {\name} to diagnose multimodal models for sentiment analysis in their research and model development. Besides the case studies, \imp{E1} mentioned \emph{``If a model is expected to achieve 90\% accuracy but is stuck at 80\%, I can use it (i.e., {\name}) to diagnose whether it is the model architecture design or the input data that limits the ultimate performance.''} 
% % The model user 
% \imp{E3} said if he finds some modalities are influential in predicting the sentiment using {\name}, he can reduce the number of modalities without losing much performance when deploying the model to low-end devices.}
% \yong{These seem not very convincing scenarios.}

\textbf{Improvements.}
% All the experts showed great interest in using {\name} to diagnose multimodal models for sentiment analysis in their research and model development.
% They 
The experts offered constructive suggestions for improvements.
\imp{E3} requested a bookmark function to save user interaction histories (\eg, selection of templates) for further review.
\imp{E1} suggested that the system can add a comparison module for exploring and comparing different models at the same time.
During the exploration, \imp{E2} and \imp{E3} observed that some large model errors are caused by dataset errors (\eg, a mismatch between the video and transcript). They recommended that the system should support correcting dataset errors.

\section{Discussion}
% In this paper, 
% In this section, 
Here, we discuss {\name} regarding generalizability, scalability, multi-level and multi-faceted exploratory analysis, and learning curve.

\textbf{Generalizability.}
{\name} was developed to visualize and explain multimodal models for sentiment analysis. We demonstrated our system through case studies on two state-of-the-art models using the CMU-MOSEI dataset. 
However, {\name} can also be used to explain other multimodal models on different sentiment datasets based on the feature importance computed by post-hoc explainability techniques.
Furthermore, the interaction types (\ie, dominance, complement, and conflict) and feature templates can summarize multimodal features from the global and subset levels in other multimodal language analyses. 
For example, for the multimodal emotion recognition task, 
the system can explain what are the dominant modalities when ``angry'' is predicted. 
The feature templates can summarize the frequent and influential feature sets for ``angry'' and facilitate the exploration of model behaviors.
% \xingbo{Currently, our system considers the widely-adopted multimodal language, facial, and acoustic features. Other features like gestures are not included.}
% \yong{Pls check it.}
% It is also interesting to incorporate other features (\eg, gestures) and examine their influences on the model predictions.

% \begin{compactitem}
% \item system can be applied to other models/multimodal sentiment datasets (regression)
% \begin{compactitem}
% \item system can be adapted to multimodal emotion classification tasks %(\eg, emotion classification)
% \end{compactitem}
% \item visual glyphs of text, audio, and vision can generally encode features about words, acoustic features, and facial expressions.
% \begin{compactitem}
% \item summary view
% \item template view
% \end{compactitem}
% \item templates, (complement, dominance, conflict)
% \end{compactitem}

\textbf{Scalability.} 
% There are also scalability issues in our approach,
Our approach also has some scalability issues,
which come from the automated algorithms and visual designs. The bottleneck of our computational cost is the feature attribution methods.
We use SHAP to compute the feature importance. It took about 25 minutes to process 2,000 instances of the CMU-MOSEI validation set.
To speed up the process, we can employ techniques such as feature clustering, data sampling, and parallel computing.
For the visual designs, the visual clutter can occur in the {\vfour}, where multimodal instances are encoded with different glyphs.
To reduce this issue, {\name} enables filtering instances according to the feature importance and sentiment predictions. 
Moreover, users can use semantic zoom to focus on instances of interest, which alleviates the overlapping issues.
% However, if there are an explosive number of instances to explore, 
% we can consider clustering them into groups and adjust the visibility of the instances in the groups with details on demand.

% \begin{compactitem}
% \item automated algorithms

% \item visual designs
% \end{compactitem}

\textbf{Multi-level and multi-faceted exploratory analysis.}
{\name} provides multi-level and multi-faceted explanations on the behaviors of multimodal models for sentiment analysis.
A general workflow for our target users (\eg, model users and researchers) starts with the {\vtwo}, where the global summary of the influences of individual modalities and their interplay is displayed. 
Then, users can specify an interaction type. Its influential and frequent multimodal features will be summarized in the {\vthree} and {\vfour}. Users can examine their error and importance patterns, which helps prioritize their efforts for the instance exploration in the {\vfive}.

\textbf{Learning curve.}
According to the feedback from the expert interviews, 
the experts pointed out that it took them some time (usually a 20-min trial) before smoothly using our system since our system contains a few components.
However, they said that {\name} is very helpful for them to explore the models. Moreover, they have derived comprehensive insights into the model behaviors and are eager to use {\name} for model understanding and diagnosis in the future.

% \textbf{Evaluation.}
% Our system was evaluated through two case studies on two models using the CMU-MOSEI dataset.
% We also collected feedback from three experts through interviews.
% In the future, we plan to 
% further evaluate our system with more multimodal models and sentiment datasets.
% Also, we would like to extend our system to other multimodal applications, such as emotion recognition.
% To fully understand the limitations and benefits of {\name}, more users should be invited to evaluate our system.

% \begin{compactitem}
% \item Invite more users to evaluate the system. test more dataset, and models
% \item Comparative study (e.g., multimodal v.s. unimodal models)
% \item \textbf{Customizable feature subsets}
% \end{compactitem}

% \yong{Where is our future work?}
% \vspace{-3mm}
\section{Conclusion and Future Work}
In this paper, we presented {\name}, a visual analytics system to help users understand and diagnose multimodal models for sentiment analysis.
{\name} provides multi-level explanations on model behaviors from language, acoustic, and visual modalities.
It features an augmented tree-like layout for a global understanding of learned intra- and inter-modal interactions. 
Moreover, the feature templates and visualization glyphs of multimodal features facilitate the exploration of a group of frequent and influential feature sets.
Through two case studies and expert interviews, we demonstrated 
% how {\name} revealed insights into 
{\name} can provide deep insights into
the state-of-art multimodal models for sentiment analysis.

In the future, we plan to enhance our system usability by adding functions, such as model comparison, data error correction. Also, we would like to extend our system to other multimodal applications (\eg, emotion recognition).
\xbRevise{Further, more domain experts can be invited to further validate the usability and effectiveness of {\name} with more datasets and models for sentiment analysis.}

% \input{sections/conclusion}

\acknowledgments{
The authors wish to thank anonymous reviewers for their feedback. 
% Yong Wang is the corresponding author. 
This research was supported in part by grant FSNH20EG01 under Foshan-HKUST Projects. }

\bibliographystyle{abbrv-doi}

\bibliography{template}
% \newpage
% \input{sections/supplementary}

\end{document}